\documentclass[10pt,twocolumn,final]{svjour3}

\usepackage[colorlinks=true,linkcolor=blue,urlcolor=blue,citecolor=blue,anchorcolor=blue,hyperfigures]{hyperref}
\usepackage[font=small,format=plain,labelfont=bf,labelsep=space,justification=justified,singlelinecheck=false]{caption}
\usepackage[numbers]{natbib}
\usepackage[misc,geometry]{ifsym} 
\usepackage[ruled,vlined,linesnumbered]{algorithm2e}
\usepackage{graphicx,url,subfig}
\graphicspath{{figs/}}
\newcommand{\examplewidth}{1.0in}
\newcommand{\shapewidth}{0.62in}
\newcommand{\figwidth}{3in}

\titlerunning{}
\authorrunning{}
\journalname{}

\begin{document}

\title{Towards the Evolution of Vertical-Axis Wind Turbines using Supershapes
}

\author{Richard J. Preen \and Larry Bull}

\institute{R. J. Preen (\Letter) \and L. Bull \at
			Department of Computer Science and Creative Technologies \\
			University of the West of England, Bristol, BS16 1QY, UK \\
			\MakeLowercase{\email{richard2.preen@uwe.ac.uk}}
		\and
		L. Bull \\ 
		\MakeLowercase{\email{larry.bull@uwe.ac.uk}}
}

\date{The final version of this paper is published in Evolutionary Intelligence 2014, 7(3):155--167.}

\maketitle

\begin{abstract}
We have recently presented an initial study of evolutionary algorithms used to design vertical-axis wind turbines (VAWTs) wherein candidate prototypes are evaluated under fan generated wind conditions after being physically instantiated by a 3D printer. That is, unlike other approaches such as computational fluid dynamics simulations, no mathematical formulations are used and no model assumptions are made. However, the representation used significantly restricted the range of morphologies explored. In this paper, we present initial explorations into the use of a simple generative encoding, known as Gielis superformula, that produces a highly flexible 3D shape representation to design VAWT. First, the target-based evolution of 3D artefacts is investigated and subsequently initial design experiments are performed wherein each VAWT candidate is physically instantiated and evaluated under fan generated wind conditions. It is shown possible to produce very closely matching designs of a number of 3D objects through the evolution of supershapes produced by Gielis superformula. Moreover, it is shown possible to use artificial physical evolution to identify novel and increasingly efficient supershape VAWT designs. 

\end{abstract}

\section{Introduction}

Renewable energy contributed over half of total net additions to global electric generating capacity from all sources in 2012, with wind power accounting for around 39\% of the renewable power added~\citep[p.~13]{Renewables:2013}. Currently, arrays of horizontal-axis wind turbines (HAWTs) are the most commonly used form of wind farm used to extract large amounts of wind energy. However, as the turbines extract the energy from the wind, the energy content decreases and the amount of turbulence increases downstream from each. For example, see \cite{Hasager:2013} for photographs and explanation of the well-known wake effect at the Horns Rev offshore wind farm in the North Sea. Due to this, HAWTs must be spaced 3--5 turbine diameters apart in the cross-wind direction and 6--10 diameters apart in the downwind direction in order to maintain 90\% of the performance of isolated HAWTs~\citep{Dabiri:2011}. The study of these wake effects is therefore a very complex and important area of research~\citep{Barthelmie:2006}, as is turbine placement~\citep{Mosetti:1994}. 

Thus, ``modern wind farms comprised of HAWTs require significant land resources to separate each wind turbine from the adjacent turbine wakes. This aerodynamic constraint limits the amount of power that can be extracted from a given wind farm footprint. The resulting inefficiency of HAWT farms is currently compensated by using taller wind turbines to access greater wind resources at high altitudes, but this solution comes at the expense of higher engineering costs and greater visual, acoustic, radar and environmental impact''~\citep[p.~1]{Dabiri:2011}. This has forced wind energy systems away from high energy demand population centres and towards remote locations with higher distribution costs. 

In contrast, vertical-axis wind turbines (VAWTs) do not need to be oriented to wind direction and the spacing constraints of HAWTs often do not apply. VAWT performance can even be increased by the exploitation of inter-turbine flow effects~\citep{Charwat:1978}. Indeed, it has recently been shown~\citep{Dabiri:2011} that power densities an order of magnitude greater can be potentially achieved by arranging VAWTs in layouts utilising counter-rotation that enable them to extract energy from adjacent wakes and from above the wind farm. 

VAWTs can also be easier to manufacture, may scale more easily, are typically inherently light-weight with little or no noise pollution, and are more able to tolerate extreme weather conditions~\citep{Eriksson:2008}. This has resulted in a recent expansion of their use in urban environments~\citep{Toja-Silva:2013}. However, their design space is complex and relatively unexplored. Generally, two classes of design are predominantly under investigation and exploitation: Savonius~\cite{Savonius:1930}, which has blades attached directly upon the central axis structure; and Darrieus~\cite{Darrieus:1931}, where the blades, either straight or curved, are positioned away from the central structure. Hybrids also exist.

We~\citep{PreenBull:2014} have recently presented an initial study of surrogate-assisted genetic algorithms (SGAs)~\citep{Dunham:1963} used to design VAWTs wherein candidate prototypes are evaluated under fan generated wind conditions after being physically instantiated by a 3D printer. That is, unlike other approaches, no mathematical formulations are used and no model assumptions are made. Initially, artificial evolution was used to explore the design space of a single isolated VAWT. Subsequently, a cooperative coevolutionary genetic algorithm (CGA)~\citep{HusbandsMill:1991} was applied to explore the design space of an array of two closely positioned VAWTs.

The results showed that EAs are capable of identifying novel and increasingly efficient VAWT designs wherein a sample of prototypes are fabricated by a 3D printer and examined for utility in the real-world. The use of a neural network surrogate model was found to reduce the number of fabrications required by an evolutionary algorithm (EA) to attain higher aerodynamic efficiency (rotation speed) of VAWT prototypes. The approach completely avoids the use of 3D computer simulations, with their associated processing costs and modelling assumptions. In particular, the wind turbine array experiment showed that surrogate-assisted coevolutionary genetic algorithms (SCGAs) are capable of iteratively increasing the performance of two closely positioned VAWTs, taking into account the inter-turbine flow effects, which is especially difficult to achieve under a high-fidelity simulation. The SCGA represents a scalable approach to the design of wind turbine arrays since the number of inputs to the surrogate-models remains constant regardless of the number of turbines undergoing evolution.

However, the representation used significantly restricted the range of morphologies explored, including a fixed number of blades. In this paper, we present initial explorations into the use of a simple generative encoding that produces a highly flexible 3D shape representation to design VAWT that are manufactured by a 3D printer and evaluated in the real world. First, the target-based evolution of 3D artefacts is investigated and subsequently initial VAWT design experiments are performed wherein each individual is physically instantiated and evaluated under fan generated wind conditions.

\section{Background}
   
\subsection{Evolving 3D shapes}

The evolution of geometric models to design arbitrary 3D morphologies has been widely explored. Early examples include Watabe and Okino's lattice deformation approach~\cite{WatabeOkino:1993} and McGuire's sequences of polygonal operators~\cite{McGuire:1993}. Sims~\cite{Sims:1994} evolved the morphology and behaviour of virtual creatures that competed in simulated 3D worlds with a directed graph encoding. Bentley~\cite{Bentley:1996} investigated the creation of 3D solid objects via the evolution of both fixed and variable length direct encodings. The objects evolved included furniture, heatsinks, penta-prisms, boat hulls, aerodynamic cars, as well as hospital department layouts. Eggenberger~\cite{Eggenberger:1997} evolved 3D multicellular organisms with differential gene expression. Jacob and Nazir~\cite{JacobNazir:2002} evolved polyhedral objects with a set of functions to manipulate the designs by adding stellating effects, shrinking, truncating, and indenting polygonal shapes. Jacob and Hushlak~\cite{JacobHushlak:2007} used an interactive evolutionary approach with L-systems~\cite{PrusinkiewiczLindenmayer:1990} to create virtual sculptures and furniture designs. 

Husbands et~al.~\cite{Husbands:1996} used an interactive evolutionary approach to design 3D objects with a superquadrics~\citep{Barr:1981} formula similar to the shape representation used here. The genetic algorithm (GA)~\cite{Holland:1975} used a directed graph encoded as bitstrings that were translated into a valid geometry. They were the first to combine superquadric primitives and global deformations with a GA, incorporating translation, rotation, scaling, reflection, tapering and twisting. A significant advantage of superquadrics is the compactness of the representation since few parameters are needed for a given deformation that widely extends the range of complex objects representable.

More recently, compositional pattern producing networks~\cite{Stanley:2007} have been used to evolve 3D objects, which were ultimately fabricated on a 3D printer~\cite{CluneLipson:2011}. Both interactive and target-based approaches were explored. Notably, Hornby et~al.~\cite{Hornby:2011} evolved and manufactured an {X}-band satellite antenna for NASA's ST5 spacecraft, representing the world's first artificially evolved hardware in space. Significantly, the evolved antennas outperformed a human design produced by the antenna contractor for the mission. Most of these approaches, however, have used simulations to provide the fitness scores of the evolved designs before final fabrication.
 
\subsection{Evolving wind turbines and blades}

The majority of blade design optimisation is performed through the use of CFD simulations, typically described with 3D Navier-Stokes equations~\citep{Anderson:1995}. However, 3D CFD simulations are computationally expensive, with a single calculation taking hours on a high-performance computer, making their use with an iterative search approach difficult~\citep{Graning:2007}. Moreover, assumptions need to be made, e.g.,\ regarding turbulence or pressure distributions, which can significantly affect accuracy. 

Previous evolutionary studies have been undertaken with types of CFD to optimise the blade profile for both HAWT~\citep{Hampsey:2002} and VAWT~\citep{Carrigan:2012} to varying degrees of success/realism. EAs have also been applied to aircraft wing design (e.g.,~\cite{OngKeane:2004}), including transonic aerofoils (e.g.,~\cite{QuagliarellaCioppa:1995,HaciogluOzkol:2003}), and multidisciplinary blade design (e.g.,~\cite{HajelaLee:1995}.) 

Most methods have used representations that define the design directly (e.g., spline surfaces~\citep{Hasenjager:2005}), or through representations designed specifically for the task (e.g., 3D aerofoils~\citep{Oyama:2001}.) Menzel and Sendhoff~\cite{MenzelSendhoff:2008} evolved parameters to a free form deformation (FFD)~\citep{SederbergParry:1986} algorithm to design the 3D stator blade of a jet turbine using CFD simulations to evaluate solutions. Instead of representing the object directly, FFD defines a lattice of control points that manipulate a given object to an arbitrary degree of complexity. As a consequence of the cost of CFD analysis, currently most blade design optimisation uses surrogate models (also known as meta models or response surface models) to reduce the number of evaluations required~\citep{Song:2002}.

A growing body of work has been exploring techniques to optimise a given wind farm layout---termed micro-siting. See \cite{Salcedo-Sanz:2011} for a recent review of evolutionary computation-based techniques. Importantly, all of this work has been based on wake models of varying degrees of fidelity; however, it has long been noted that as the interaction dynamics become more complex, simulation deficiencies can severely mislead the evolutionary search, leading to a `reality gap' once the design is physically instantiated~\citep{Jakobi:1995}.

\subsection{Evolving physical artefacts}

The evaluation of physical artefacts directly for fitness determination can be traced back to the origins of evolutionary computation~\citep{Dunham:1963}. For example, the first evolution strategies were used to design jet nozzles with a string of section diameters, which were then machined and tested for fitness~\citep{Rechenberg:1971}. Other well-known examples include robot controller design~\citep{Nolfi:1992}, electronic circuit design using programmable hardware~\citep{Thompson:1998}, product design via human provided fitness values~\citep{Herdy:1996}, chemical systems~\citep{Theis:2006}, and unconventional computers~\citep{HardingMiller:2004}. More recently, \cite{Boria:2009} used an EA to evolve a morphing wing structure where physical designs were morphed using a set of actuators and evaluated in a closed-loop wind tunnel. 

Evolution in hardware has the potential to benefit from access to a richer environment where it can exploit subtle interactions that can be utilised in unexpected ways. For example, the EA used by Thompson~\citep{Thompson:1998} to work with field-programmable gate array circuits used physical properties of the system to solve problems where the properties used are still not understood. Humans can be prevented from designing systems that exploit these subtle and complex physical characteristics through a lack of understanding, however this does not prevent exploitation through artificial evolution. There is thus a real possibility that evolution in hardware may allow the discovery of new physical effects, which can be harnessed for computation/optimisation~\citep{MillerDowning:2002}.  

Moreover, the advent of high quality, low-cost, additive rapid fabrication technology (known as 3D printing) means it is now possible to fabricate a wide range of prototype designs quickly and cheaply. 3D printers are now capable of printing an ever growing array of different materials, including food, e.g.,\ chocolate~\citep{Hao:2009} and meat~\citep{Lipton:2010} for culinary design; sugar, e.g.,\ to help create synthetic livers~\citep{Miller:2012}; chemicals, e.g.,\ for custom drug design~\citep{Cronin:2012}; cells, e.g.,\ for functional blood vessels~\citep{Jakeb:2008} and artificial cartilage~\citep{Xu:2013}; plastic, e.g.,\ Southampton University laser sintered aircraft; thermoplastic, e.g.,\ for electronic sensors~\citep{Leigh:2012}; titanium, e.g.,\ for prosthetics such as the synthetic mandible developed by the University of Hasselt and transplanted into an 83-year old woman; and liquid metal, e.g.,\ for stretchable electronics~\citep{Ladd:2013}. One potential benefit of the technology is the ability to perform fabrication directly in the target environment; for example, Cohen et~al.\@~\cite{Cohen:2010} recently used a 3D printer to perform a minimally invasive repair of the cartilage and bone of a calf femur {\it in situ}. Lipson and Pollack~\cite{LipsonPollack:2000} were the first to exploit the emerging technology in conjunction with an EA using a simulation of the mechanics and control, ultimately printing mobile robots with embodied neural network controllers.

Funes and Pollack~\cite{FunesPollack:1998} performed one of the earliest attempts to physically instantiate evolved 3D designs by placing physical {LEGO} bricks according to the schematics of the evolved individuals. A direct encoding of the physical locations of the bricks was used and the fitness was scored using a simulator which predicted the stability of the composed structures. Additionally, Hornby and Pollack~\cite{HornbyPollack:2001} used L-systems to evolve furniture designs, which were then manufactured by a 3D printer. They found the generative encoding of L-systems produced designs faster and with higher fitness than a non-generative system. Generative systems are known to produce more compact encodings of solutions and thereby greater scalability than direct approaches (e.g.,\ see~\cite{Schoenauer:1996}).

Recently, Rieffel and Sayles~\cite{RieffelSayles:2010} evolved circular 2D shapes where each design was fabricated on a 3D printer before assigning fitness values. Interactive evolution was undertaken wherein the fitness for each printed shape was scored subjectively. Each individual's genotype consisted of twenty linear instructions which directed the printer to perform discrete movements and extrude the material. As a consequence of performing the fitness evaluation in the environment, that is, after manufacture, the system as a whole can exhibit epigenetic traits, where phenotypic characteristics arise from the mechanics of assembly. One such example was found when selecting shapes that most closely resembled the letter `A'. In certain individuals, the cross of the pattern was produced from the print head dragging a thread of material as it moved between different print regions and was not explicitly instructed to do so by the genotype.

\section{Gielis superformula}

Superquadrics~\citep{Barr:1981} have long been used for modelling 3D objects, including the more recent computational simplification, ratioquadrics~\citep{BlancSchlick:1996}. However, superquadrics are limited by an intrinsic symmetry. Extensions through additional local and global deformations have therefore been proposed~\citep{Barr:1984}, with such deformations requiring larger sets of parameters. Global deformations affect the whole superquadric and include tapering, bending, twisting, or any hierarchical combination thereof. Other methods of increasing the degrees of freedom include using Bezier curves as functions in the exponent of superquadric equations~\citep{ZhouKambhamettu:1999} and hyperquadrics~\citep{Hanson:1988}. 

Despite the proposed extensions to superquadrics, they are fixed to the orthogonal system of coordinate axes. It is therefore difficult to describe certain shapes such as polygons or polyhedrons with superquadrics. Gielis~\cite{Gielis:2003b} therefore introduced a new approach using a generalised superellipse equation, termed the {\em superformula}, which can be defined for any symmetry. Modifying the set of real-valued parameters to the superformula generates {\em supershapes}, myriad and diverse natural polygons with corresponding degrees of freedom. The superformula can be used to create 3D objects using the spherical product of two superformulas. 

``In general, one could think of the basic superformula as a transformation to fold or unfold a system of orthogonal coordinate axes like a fan. This creates a basic symmetry and metrics in which distances can further be deformed by local or global transformations. Such additional transformations increase the plasticity of basic supershapes''~\cite[p.~337]{Gielis:2003b}. Gielis superformula can be further generalised to increase the degrees of freedom, adding twist and additional rotations, permitting the creation of more complex 3D forms, including shells, m\"obius strips, and umbilic tori. 

Gielis superformula, which defines a supershape in 2D is given in the following equation, where $r$ is the radius; $\phi$ is the angle; $a>0$, $b>0$ control the size of the supershape and typically = 1; and $m$ (symmetry number), $n_1$, $n_2$ and $n_3$ (shape coefficients) are the real-valued parameters:
\begin{equation}
	r=f(\phi)\frac{1}{\sqrt[n_1]{(|\frac{1}{a}cos(\frac{m}{4} \phi)|)^{n_2} + (|\frac{1}{b}sin(\frac{m}{4} \phi)|)^{n_3}}}
\end{equation}
Using the spherical product, the basic extension to 3D:
\begin{equation}
	x = r_1(\theta) \times cos(\theta) \times r_2(\varphi) \times cos(\varphi)
\end{equation}
\begin{equation}
	y = r_1(\theta) \times sin(\theta) \times r_2(\varphi) \times cos(\varphi)
\end{equation}
\begin{equation}
	z = r_2(\varphi) \times sin(\varphi)
\end{equation}
Where $-\frac{\pi}{2} \le \varphi \le \frac{\pi}{2}$ for latitude and $-\pi \le \theta \le \pi$ for longitude.

More complex shapes can be produced through additional deformation parameters that apply toroidal instead of spherical mappings, alter the radius and diameter of the toroid, modify the offset along the rotation axis, and perform additional twist and further rotations; procedures to calculate the extended superformula are shown in Algorithm~\ref{alg:superformula}.
 
\begin{algorithm}[t]
	\SetAlgoLined\DontPrintSemicolon
	\SetKwProg{myalg}{Algorithm}{}{}
	\SetKwProg{myproc}{Procedure}{}{}

	\SetKwFunction{superformula}{Superformula}
	\SetKwFunction{extendedsuperformula}{Extended\_Superformula}

	\myproc{\extendedsuperformula{$\theta$, $\phi$, $m_1$, $n_{1,1}$, $n_{1,2}$, $n_{1,3}$, $m_2$, $n_{2,1}$, $n_{2,2}$, $n_{2,3}$, $c_1$, $c_2$, $c_3$, $t_1$, $t_2$, $d_1$, $d_2$, $r_o$}}{
		$t_{2c} \leftarrow (r_0 \times c_2^{d_2} \times t_2 \times c_1) / 2$ \;
		$t_2 \leftarrow t_2 \times c_1 \times \theta$ \;
		$d_1 \leftarrow (\theta \times c_1)^{d_1}$ \;
		$d_2 \leftarrow (\theta \times c_2)^{d_2}$ \;
		$\theta \leftarrow (((\pi \times 2) \times \theta) - \pi) \times c_1$ \;
		$\phi \leftarrow ((\pi \times \phi) - (\pi / 2)) \times c_2$ \;
		$\phi_2 \leftarrow \phi + (c_2 \times \theta)$ \;
		$r_1 \leftarrow$ Superformula($\theta, 1, 1, m_1, n_{1,1}, n_{1,2}, n_{1,3}$) \;
		$r_2 \leftarrow$ Superformula($\phi, 1, 1, m_2, n_{2,1}, n_{2,2}, n_{2,3}$) \;
		$x \leftarrow r_0 \times r_1 \times (t_1 + d_1 \times r_2 \times cos(\phi_2)) \times cos(\phi)$ \;
		$y \leftarrow r_0 \times r_1 \times (t_1 + d_1 \times r_2 \times cos(\phi_2)) \times sin(\phi)$ \;
		$z \leftarrow r_0 \times d_2 \times (r_2 \times sin(\phi_2) - t_2) + t_{2c}$ \;
		\KwRet $x$, $y$, $z$ \;
	}

	\nl

	\myproc{\superformula{$\phi$, $a$, $b$, $m$, $n_1$, $n_2$, $n_3$}}{
		\KwRet $((|cos((m \times \phi)/4)/a|)^{n_2} + (|sin((m \times \phi)/4)/b|)^{n_3})^{-1/n_1}$ \;
	}
	\caption{Extended superformula.}
	\label{alg:superformula}
\end{algorithm} 
 
\begin{figure}[t]
	\centering
	\subfloat[Cube $m_1=4$, $n_{1,1}=10$, $n_{1,2}=10$, $n_{1,3}=10$, $m_2=4$, $n_{2,1}=10$, $n_{2,2}=10$, $n_{2,3}=10$] {\label{fig:cube}\includegraphics[width=\examplewidth]{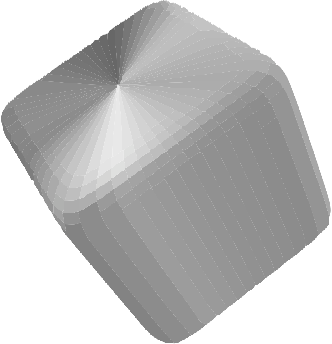}}
	\hspace{0.5in}
	\subfloat[Star $m_1=6$, $n_{1,1}=5$, $n_{1,2}=10$, $n_{1,3}=10$, $m_2=4$, $n_{2,1}=10$, $n_{2,2}=10$, $n_{2,3}=10$] {\label{fig:star}\includegraphics[width=\examplewidth]{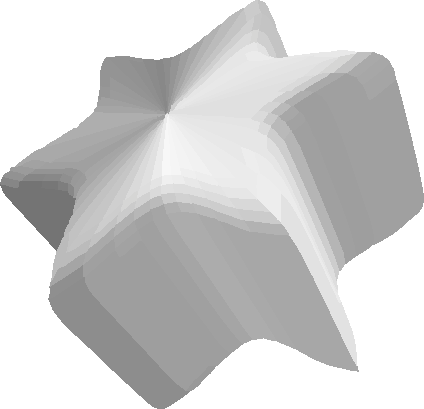}}\\
	\subfloat[Heart $m_1=3$, $n_{1,1}=1.5$, $n_{1,2}=12$, $n_{1,3}=3$, $m_2=0$, $n_{2,1}=3$, $n_{2,2}=0$, $n_{2,3}=0$] {\label{fig:heart}\includegraphics[width=\examplewidth]{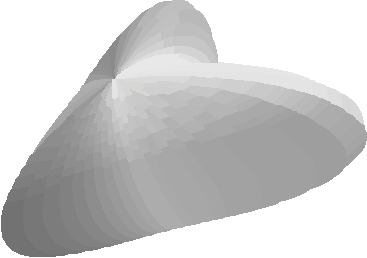}}
	\hspace{0.5in}
	\subfloat[Shell $m_1=3$, $n_{1,1}=1.5$, $n_{1,2}=12$, $n_{1,3}=3$, $m_2=0$, $n_{21}=3$, $n_{22}=0$, $n_{23}=0$, $t_2=2$, $d_1=1$, $d_2=1$, $c_1=5$] {\label{fig:shell}\includegraphics[width=\examplewidth]{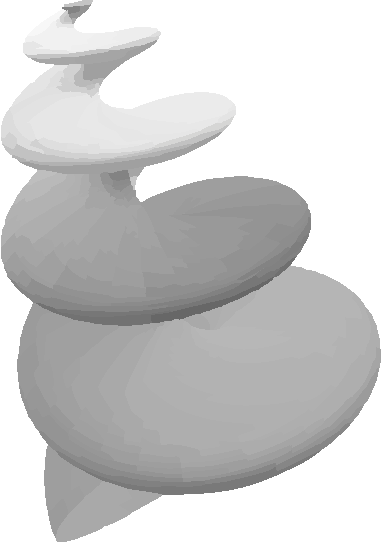}}\\
	\subfloat[Torus $m_1=10$, $n_{1,1}=10$, $n_{1,2}=10$, $n_{1,3}=10$, $m_2=10$, $n_{2,1}=10$, $n_{2,2}=10$, $n_{2,3}=10$ $t_1=2$, $c_3=0$]{\label{fig:torus}\includegraphics[width=\examplewidth]{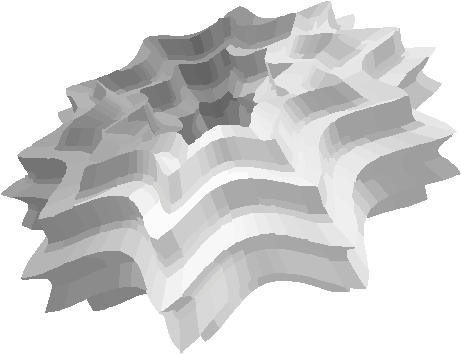}}
	\hspace{0.5in}
	\subfloat[M\"obius Strip $m_1=3$, $n_{1,1}=1.5$, $n_{1,2}=12$, $n_{1,3}=3$, $m_2=0$, $n_{2,1}=3$, $n_{2,2}=0$, $n_{2,3}=0$ $t_1=4$, $t_2=0$, $d_1=0$, $d_2=0$, $c_1=5$, $c_2=0.3$, $c_3=2.2$]{\label{fig:mobius}\includegraphics[width=\examplewidth]{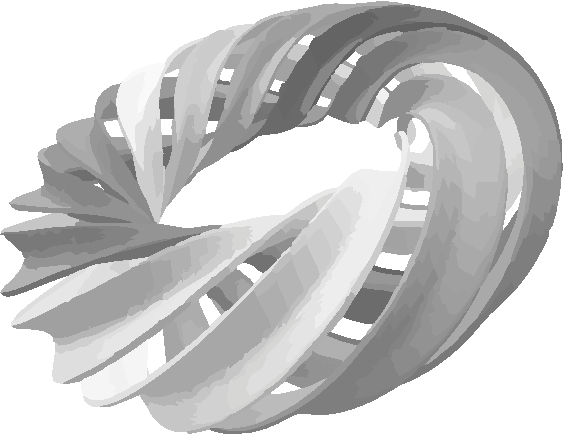}}
	\caption{Example 3D supershapes.}
	\label{fig:superformula-shapes}
\end{figure}

Example shapes\footnote{A supershape visualisation tool and its source code produced by Martin Schneider, licensed under Creative Commons Attribution Share Alike 3.0 and GNU GPL license, can be found at \url{http://openprocessing.org/visuals/?visualID=2638}.} generated with the superformula can be seen in Fig.~\ref{fig:superformula-shapes} where the cube, star, and heart can be generated from the same set of eight real-valued parameters; the torus requiring two additional parameters; the shell a total of twelve; and the m\"obius strip a total of fifteen.

In contrast to the much studied superquadrics, there is relatively little prior work exploring applications of Gielis superformula. Morales et~al.~\cite{Morales:2008} used a GA to evolve $N$-dimensional superformula for clustering. Several 3D supershapes with human specified superformula parameters have been used to represent dielectric antennas that were evaluated under numerical simulations~\citep{Simeoni:2009,Simeoni:2011,Bia:2013}, leading to the January 2013 commercialisation of the representation for ultra-wideband antennas by Antenna Company,\footnote{Antenna Company \url{http://www.antennacompany.com}} Willemstad, Cura\c{c}ao, Kingdom of the Netherlands.

Given a target shape it is often very useful to identify a representative formula, e.g., for compression. Optimisation methods, such as the Levenberg-Marquardt (LM) theory~\cite{Press:1992}, have typically been used to identify the best fitting superquadric parameters (e.g., \cite{GuptaBajcsy:1993}). However LM cannot retrieve all of the parameters required for supershape fitting. Bokhabrine et~al.~\cite{Bokhabrine:2007} used a GA to evolve all supershape parameters for surface reconstruction (i.e., a target-based approach) using an inside-outside function~\cite{Fougerolle:2006} for fitness. Voisin et~al.~\cite{Voisin:2009} later extended this to utilise a pseudo-Euclidean distance for fitness determination, yielding improved performance.

\section{Target-based evolution}

The ability to inject complex geometric patterns into an EA for further optimisation is an important area of research~\citep{Clune:2013}. The target-based evolution of a given design is therefore initially explored here as this can provide a simple mechanism to seed an initial population used for physical evolution. This may allow the future seeding of an EA with human designed VAWTs represented as Gielis superformula, thus speeding up the search for more efficient solutions.

The cube, star, and heart shapes (as seen in Fig.~\ref{fig:superformula-shapes}) are here converted into $50\times50\times50$ binary voxel arrays and used as the desired targets, where the fitness of an individual is the fraction of voxels that correctly match. The genotype of each individual in the population consists of eight real-valued parameters in the range [0,50] which affect the superformula, giving rise to the supershape. The GA proceeds with a population, $P$, of 200 individuals, a per allele mutation rate of 25\%, and mutation step size of $\pm[0,5]$; a crossover rate of 0\%; the GA tournament size for both selection and replacement is set to 3.

\begin{figure}[t]
	\centering 
	\includegraphics[width=3.3in]{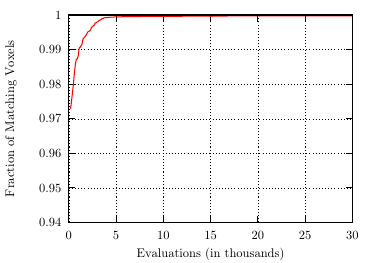}
	\caption{Evolution of a 3D cube.}
	\label{fig:ga-cube}
	\vspace{0.2cm}
	\includegraphics[width=3.3in]{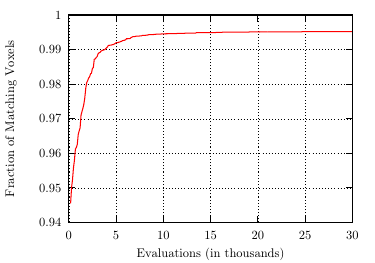}
	\caption{Evolution of a 3D star.}
	\label{fig:ga-star}
	\vspace{0.2cm}
	\includegraphics[width=3.3in]{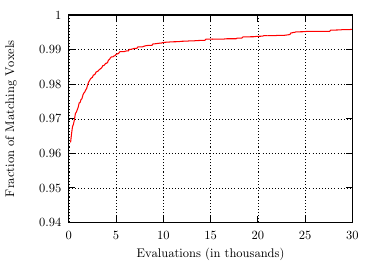}
	\caption{Evolution of a 3D heart.}
	\label{fig:ga-heart}
\end{figure}

Figs.~\ref{fig:ga-cube}--\ref{fig:ga-heart} show the fraction of total voxels matched to the target shapes during evolution of the supershapes; results presented are an average of 10 experiments. Similar to \cite{CluneLipson:2011}, a large number of voxels are quickly matched, however here the target object is not identifiable until approximately 99\% are set correctly. As such, the small differences in fitness between the individuals represent substantial differences in whether the target object is recognisable. In all cases, greater than 99.5\% fitness is achieved. From Fig.~\ref{fig:ga-cube} it can be seen that, on average, the GA takes approximately 1100 evaluations to reach $>$99\% matching voxels of a target cube object and 3700 evaluations to achieve $>$99.9\%. Fig.~\ref{fig:ga-star} shows that on average approximately 3900 evaluations are required to reach $>$99\% matching voxels of a target star object and 16100 evaluations to achieve $>$99.5\%. Finally, Fig.~\ref{fig:ga-heart} shows that, on average, $>$99\% matching voxels of a target heart object is reached after 6400 evaluations and $>$99.5\% after 24000 evaluations. 

\begin{figure}[t]
	\centering 
	\subfloat[96.71\%]{\label{fig:cube-9671}\includegraphics[width=\shapewidth]{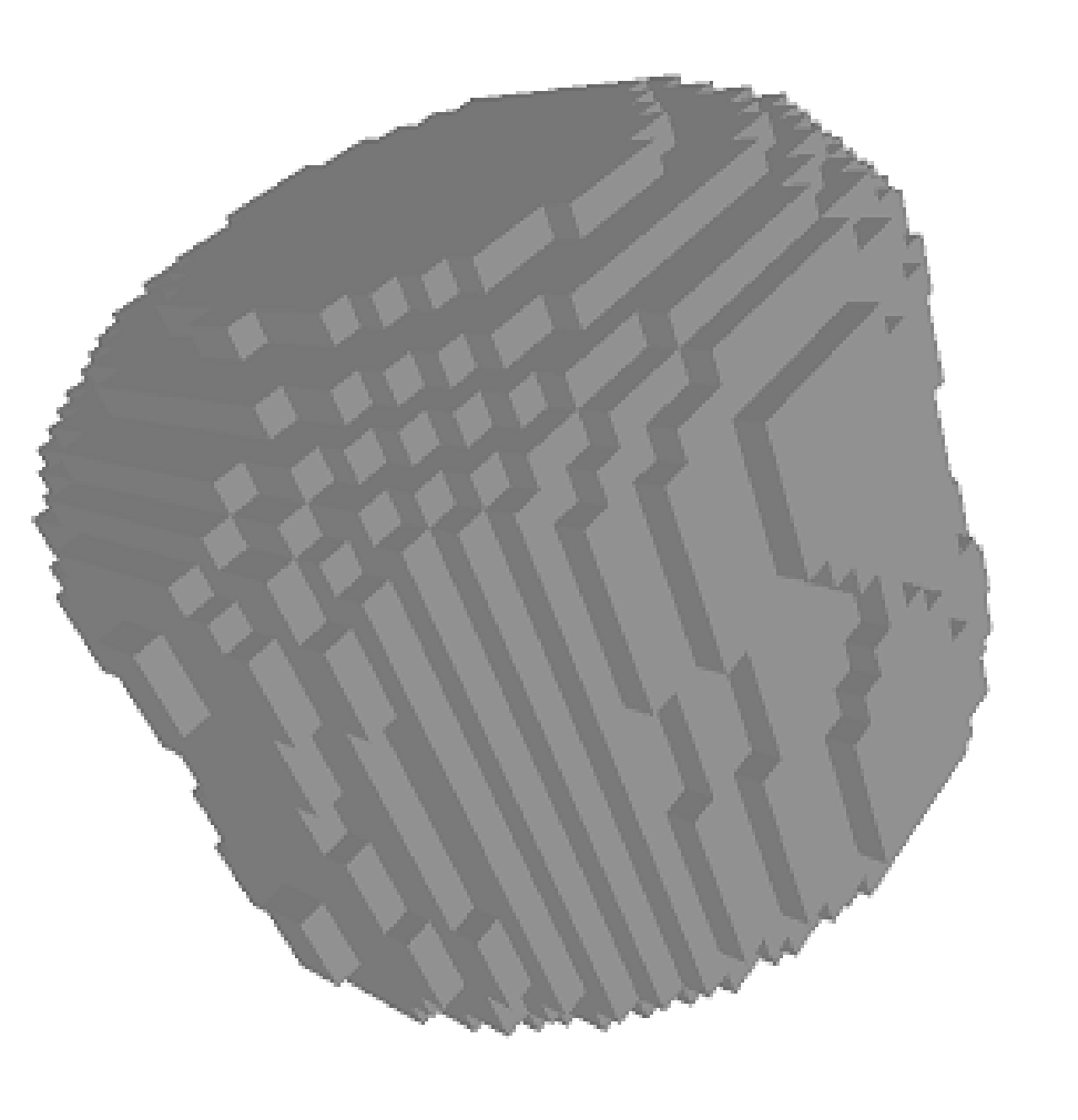}}
   	\subfloat[98.50\%]{\label{fig:cube-9850}\includegraphics[width=\shapewidth]{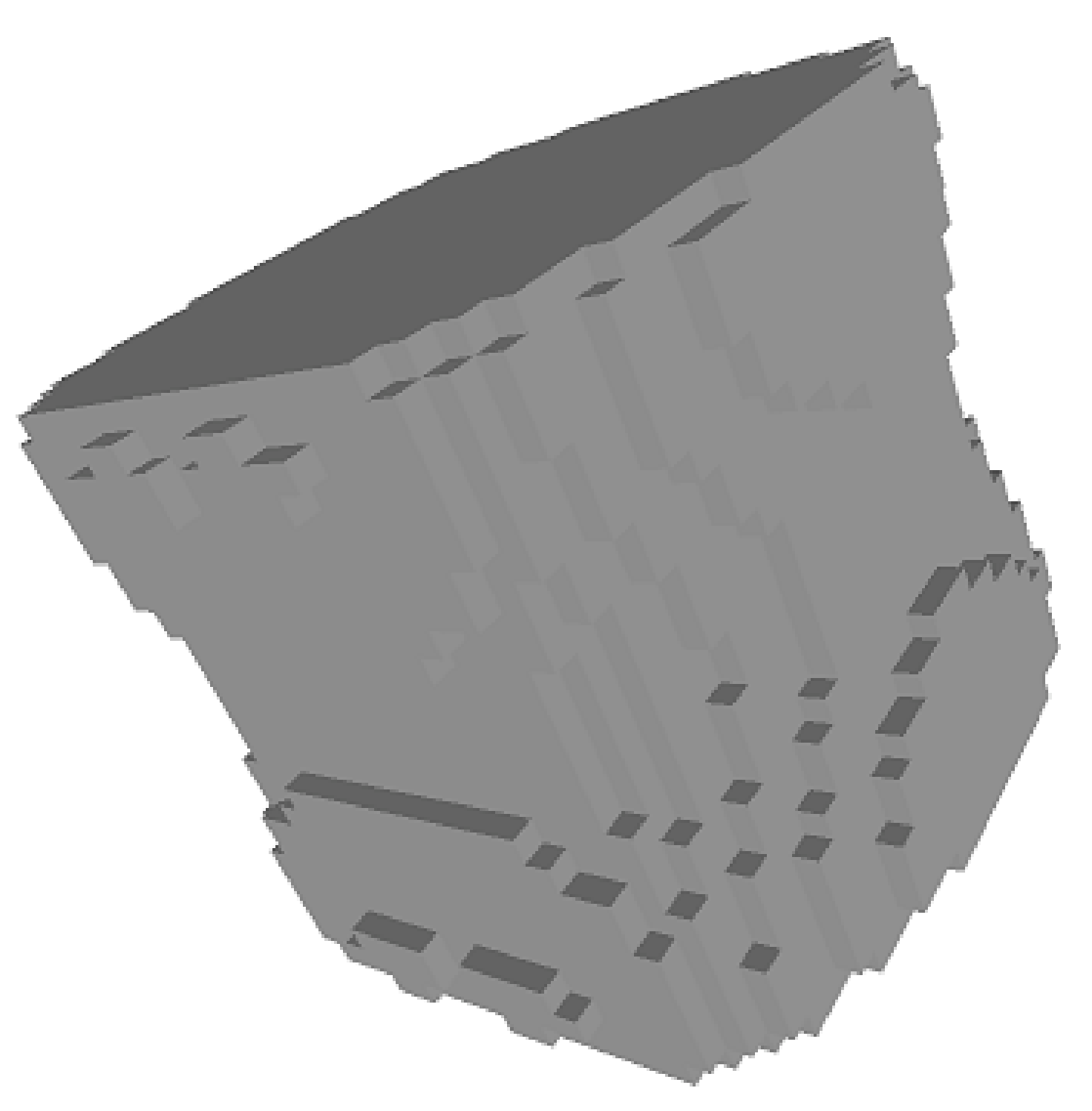}}
   	\subfloat[99.29\%]{\label{fig:cube-9929}\includegraphics[width=\shapewidth]{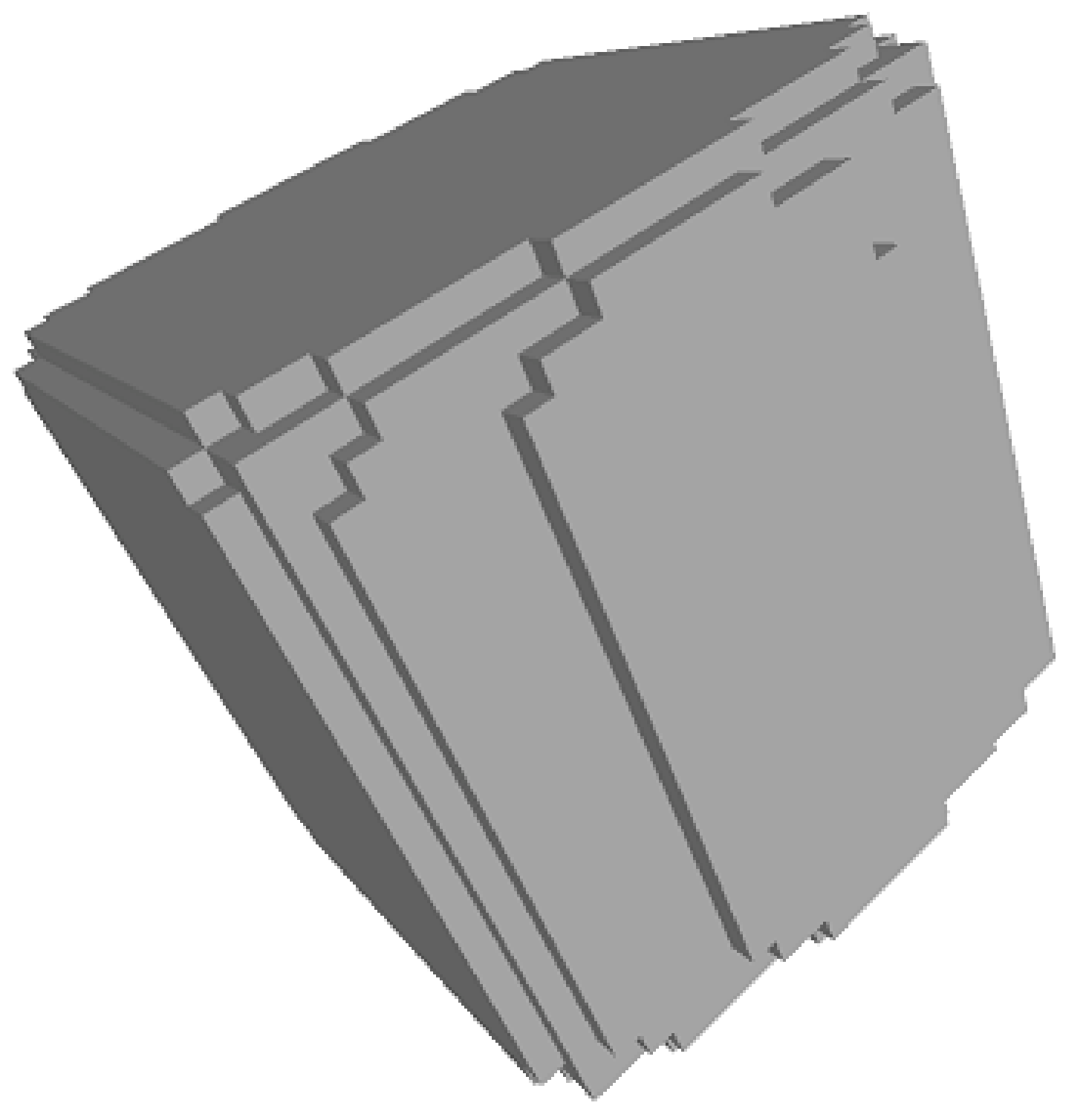}}
   	\subfloat[99.97\%]{\label{fig:cube-9997}\includegraphics[width=\shapewidth]{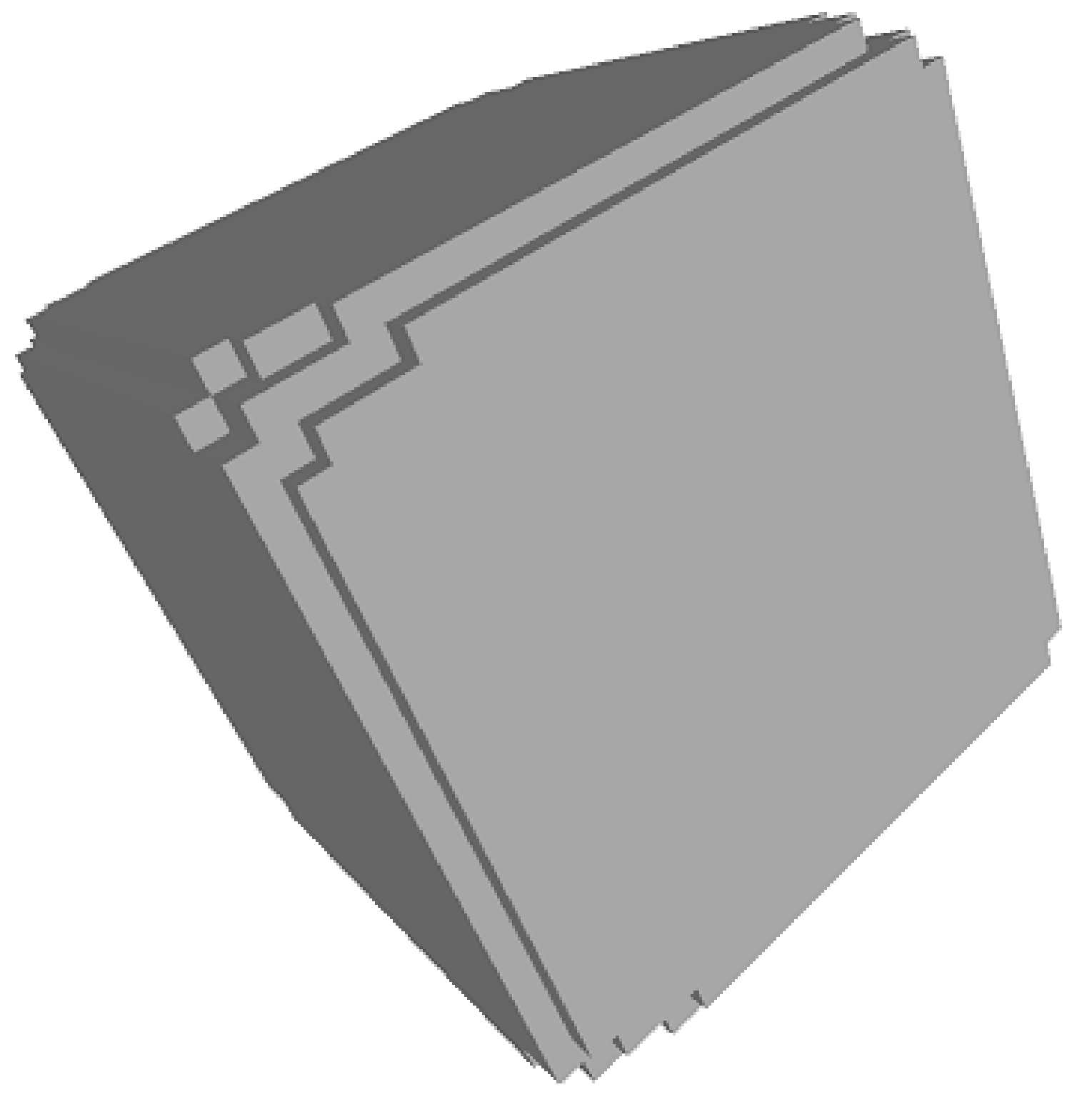}}
   	\subfloat[Target]{\label{fig:cube-1000}\includegraphics[width=\shapewidth]{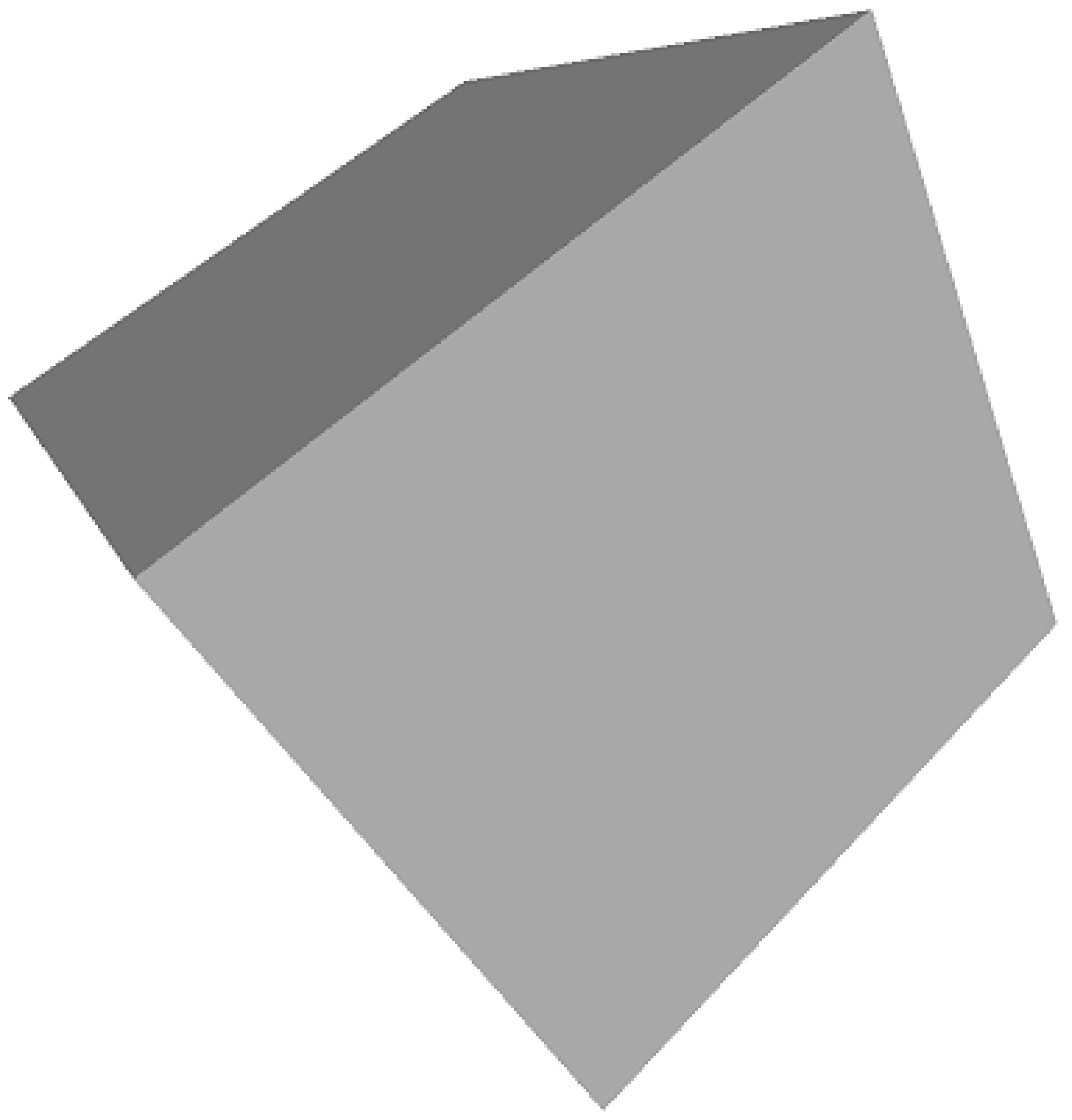}}
   	\caption{Evolution of a 3D cube.}
   	\label{fig:cube-evolution}
\end{figure}

\begin{figure}[t]
	\centering 
	\subfloat[94.18\%]{\label{fig:star-9418}\includegraphics[width=\shapewidth]{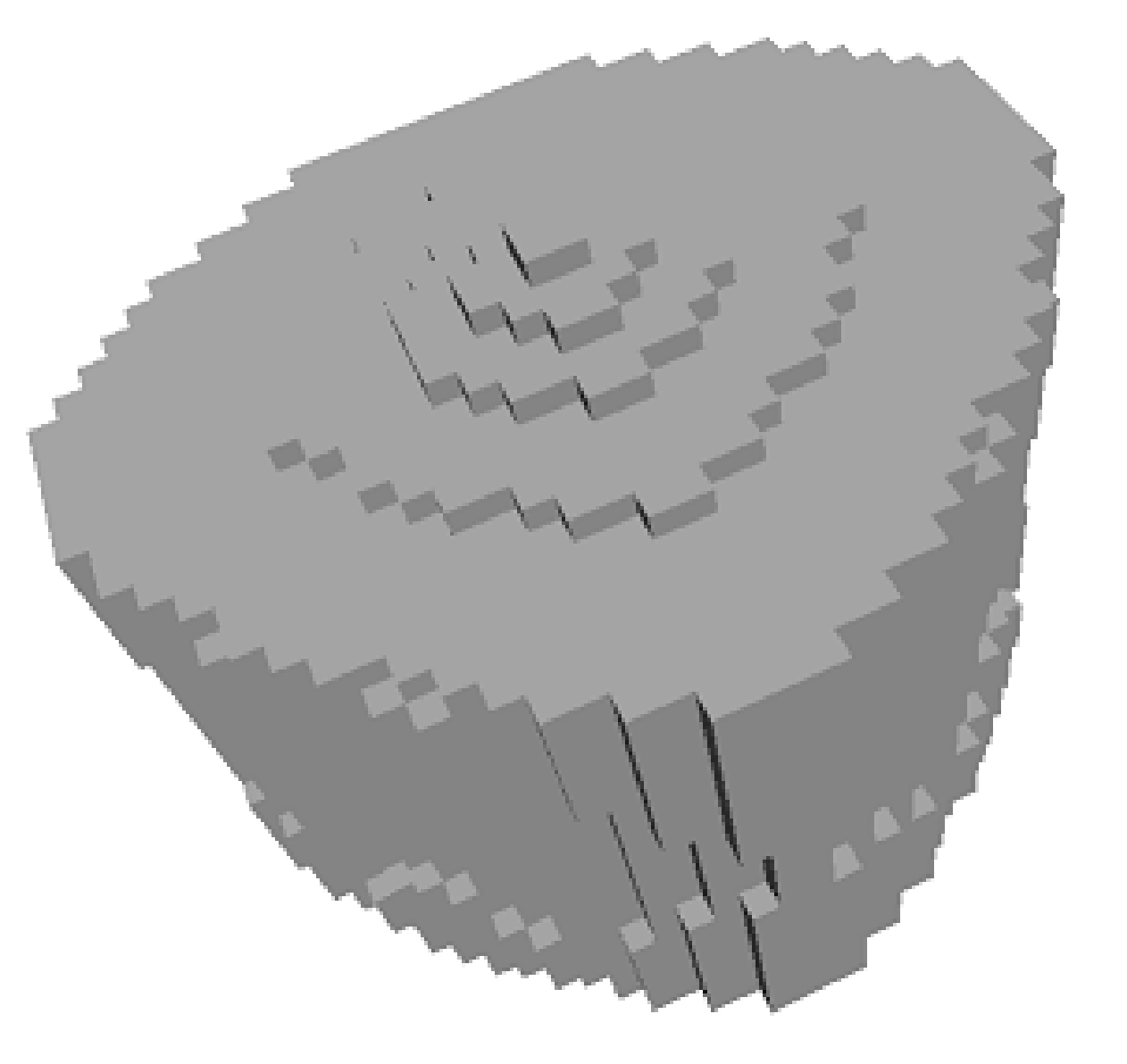}}
	\subfloat[96.24\%]{\label{fig:star-9624}\includegraphics[width=\shapewidth]{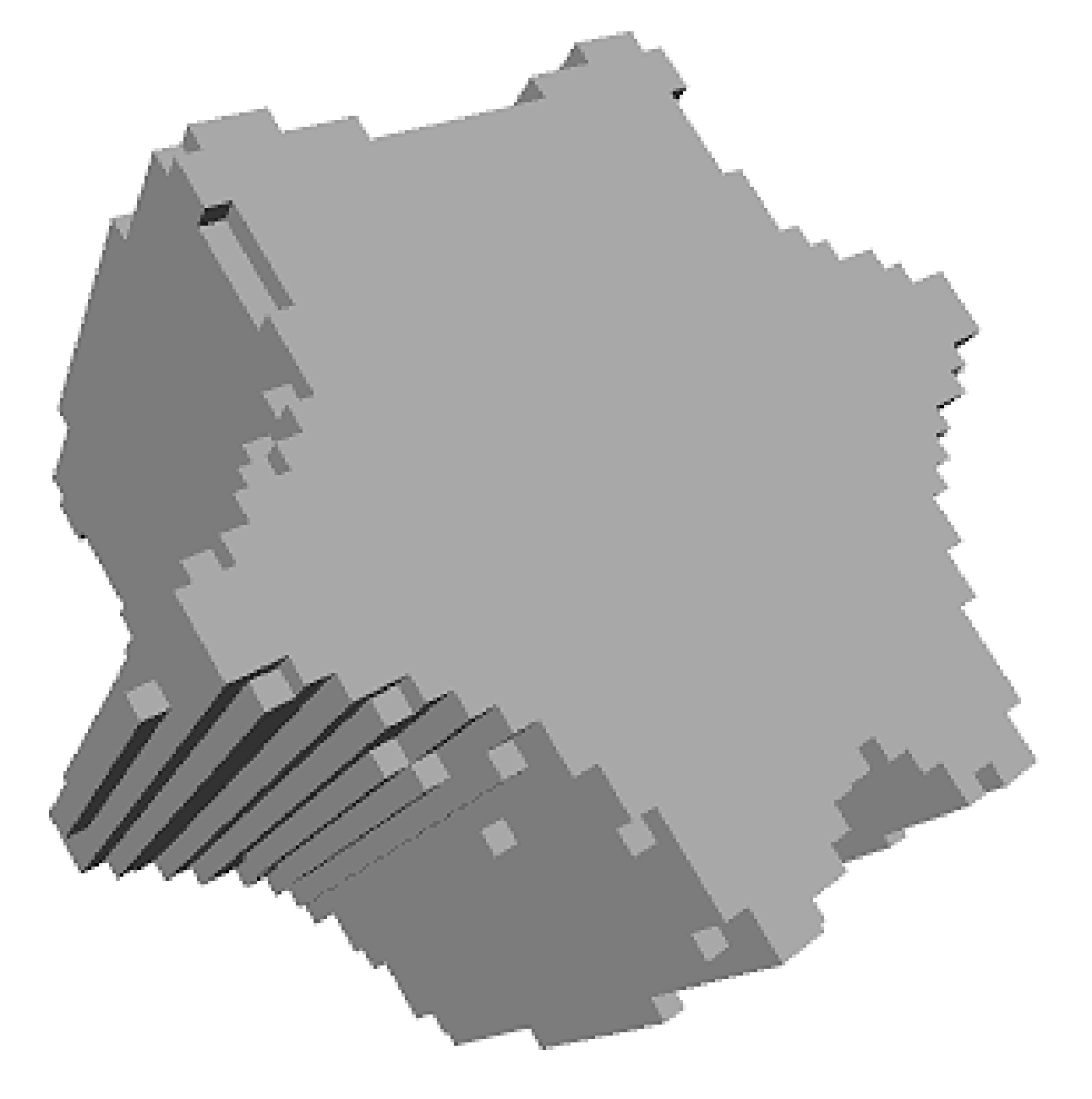}}
	\subfloat[98.03\%]{\label{fig:star-9803}\includegraphics[width=\shapewidth]{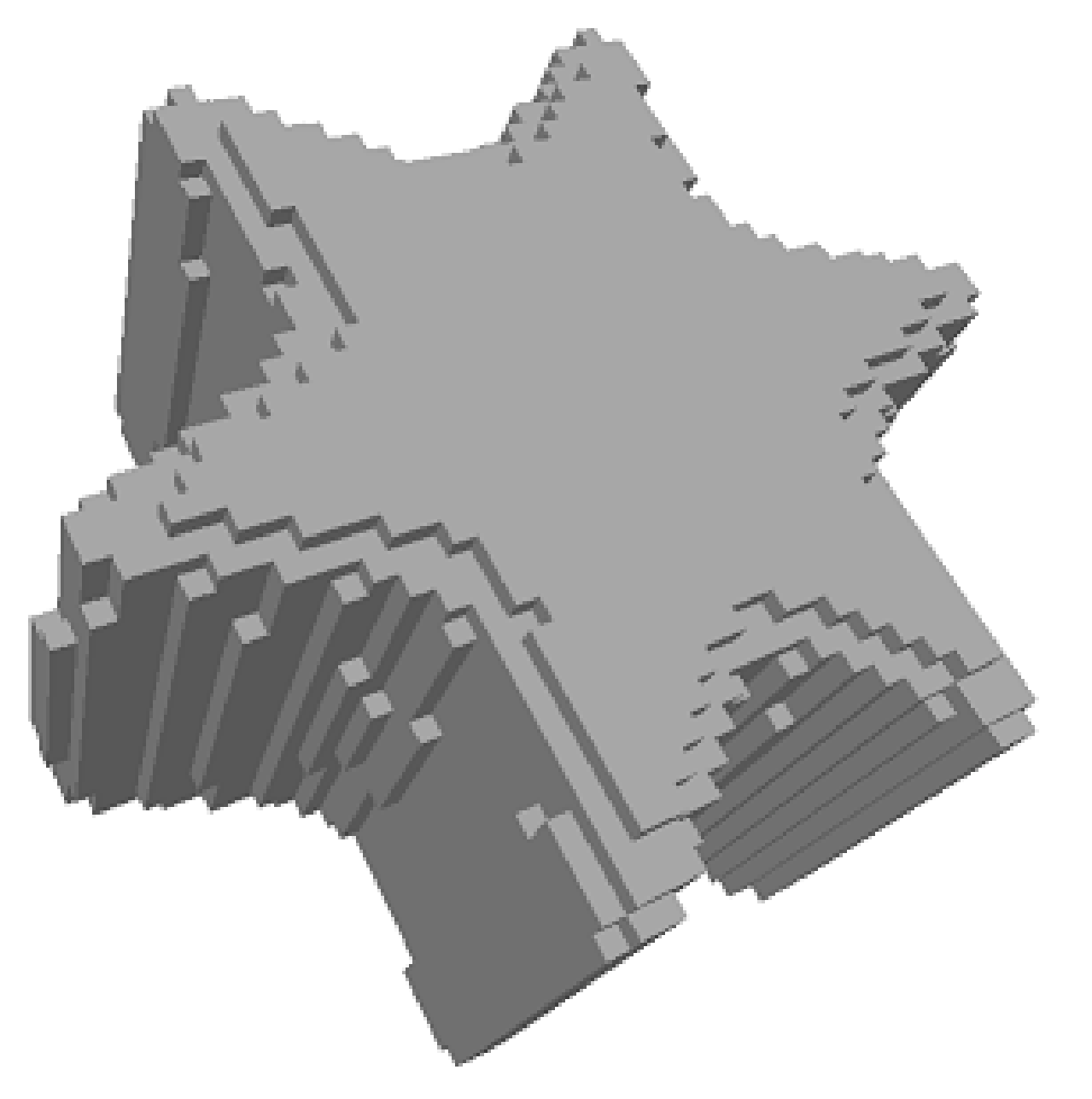}}
	\subfloat[99.56\%]{\label{fig:star-9956}\includegraphics[width=\shapewidth]{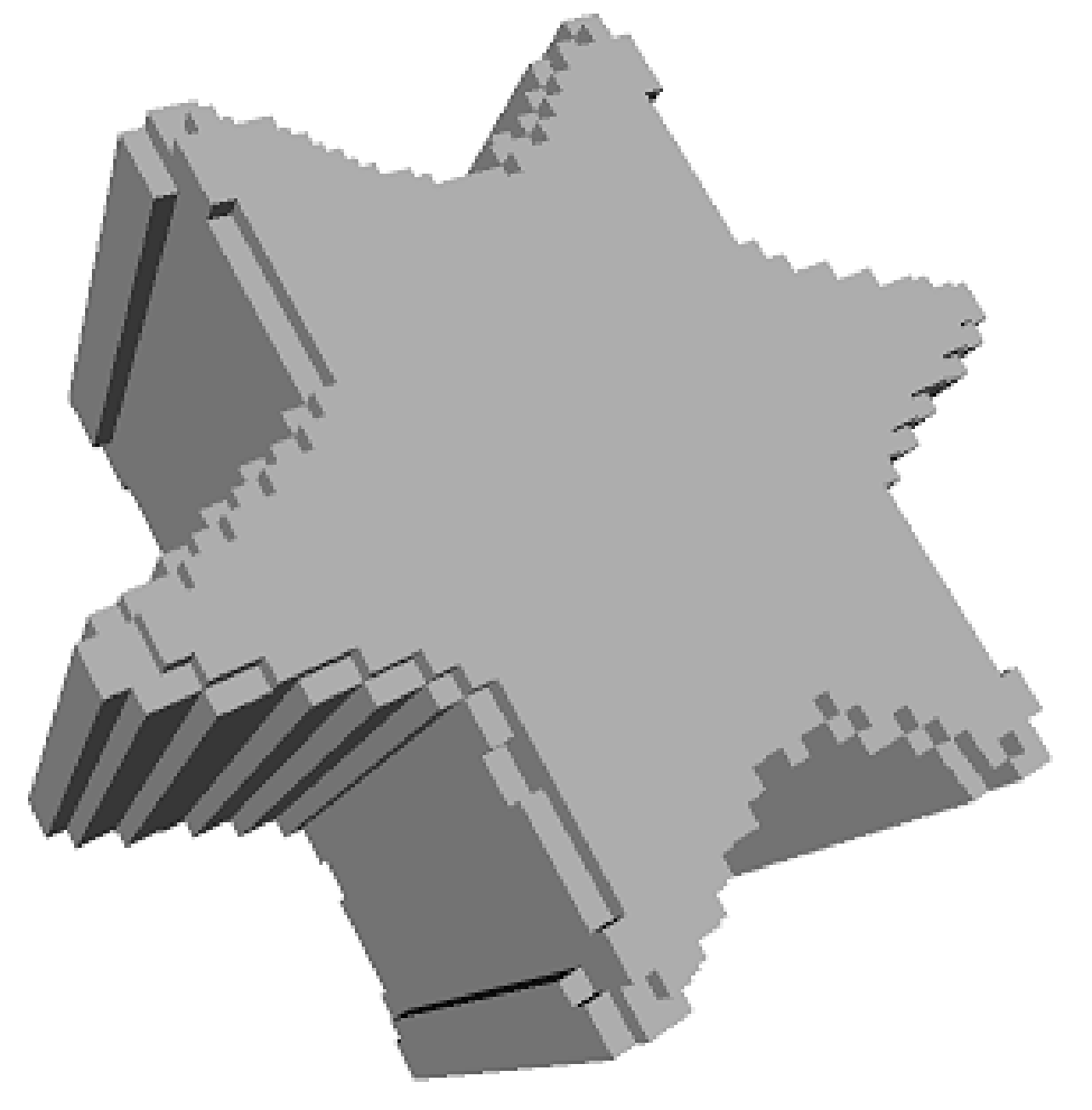}}
	\subfloat[Target]{\label{fig:star-1000}\includegraphics[width=\shapewidth]{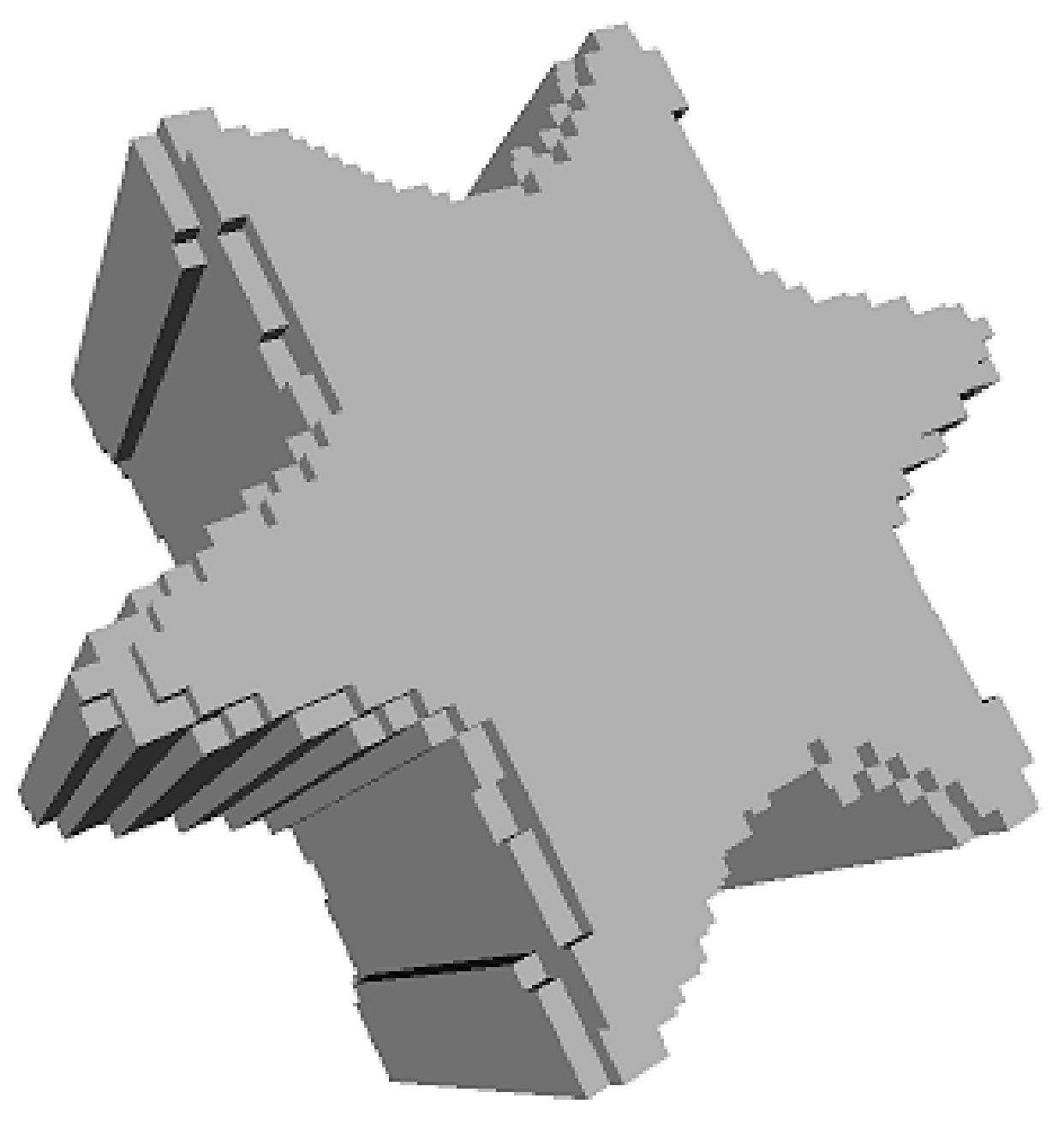}}
	\caption{Evolution of a 3D star.}
	\label{fig:star-evolution}
\end{figure}

\begin{figure}[t]
	\centering 
	\subfloat[96.11\%]{\label{fig:heart-9611}\includegraphics[width=\shapewidth]{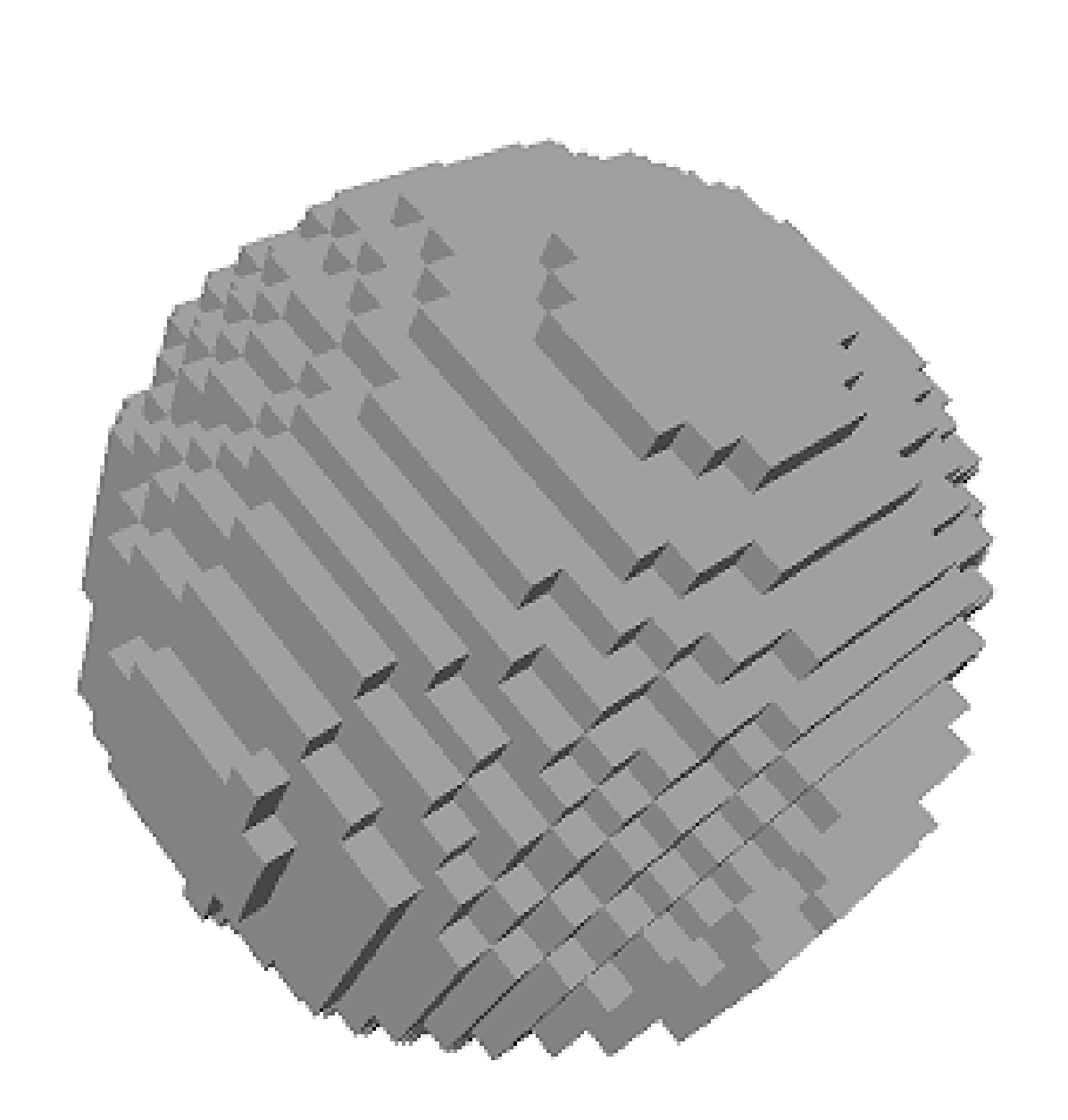}}
	\subfloat[98.11\%]{\label{fig:heart-9811}\includegraphics[width=\shapewidth]{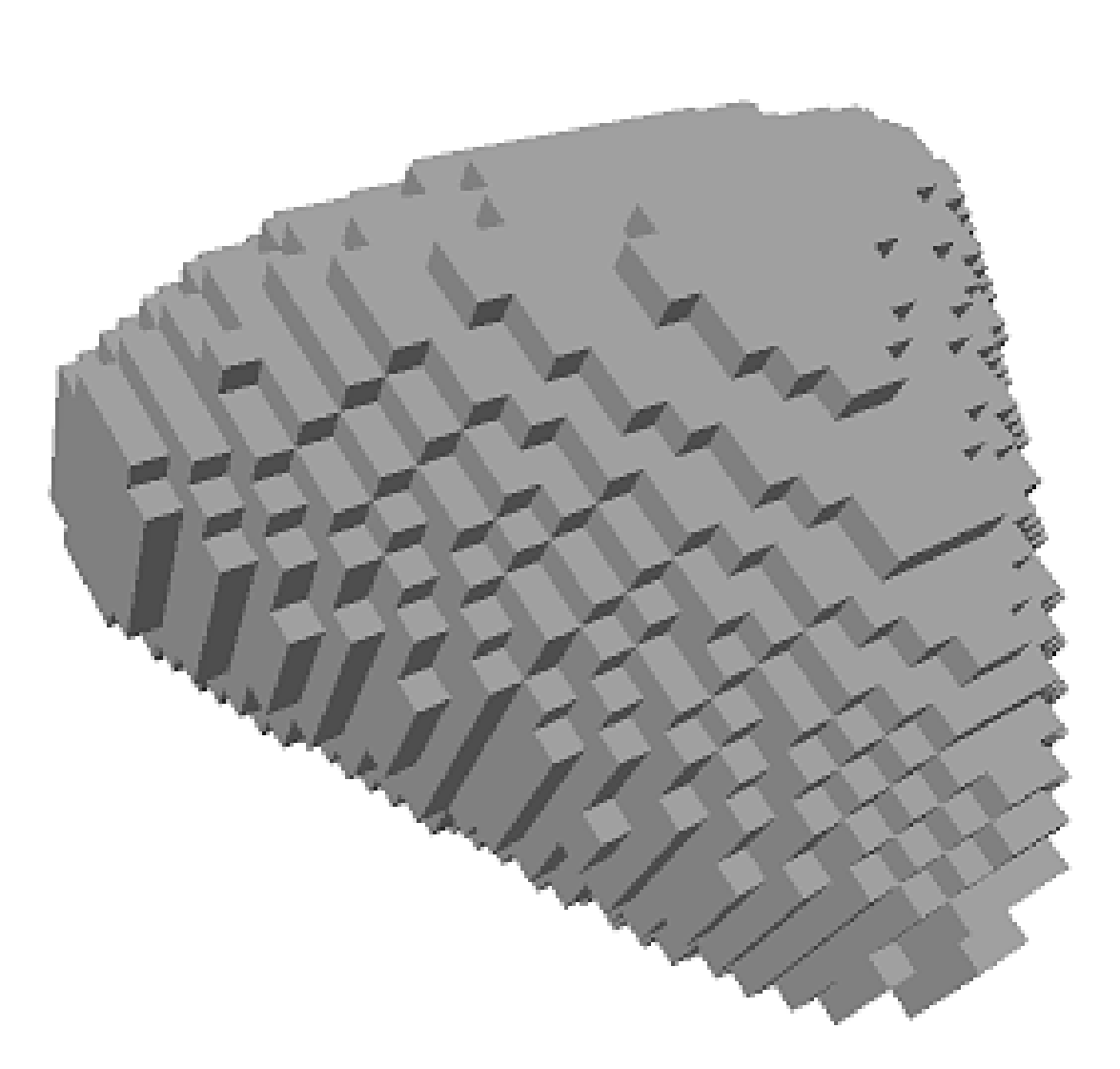}}
	\subfloat[99.00\%]{\label{fig:heart-9900}\includegraphics[width=\shapewidth]{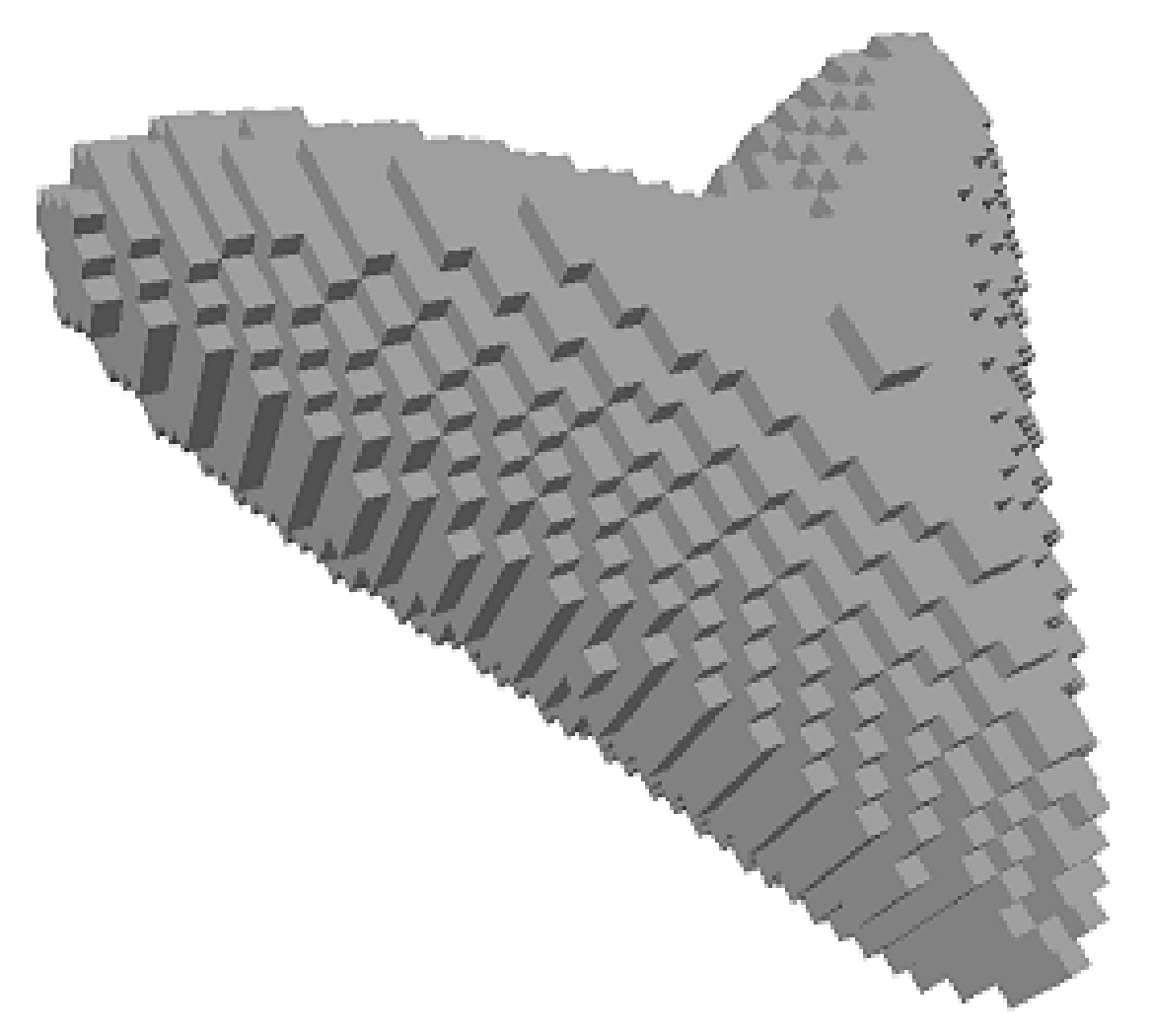}}
	\subfloat[99.50\%]{\label{fig:heart-9950}\includegraphics[width=\shapewidth]{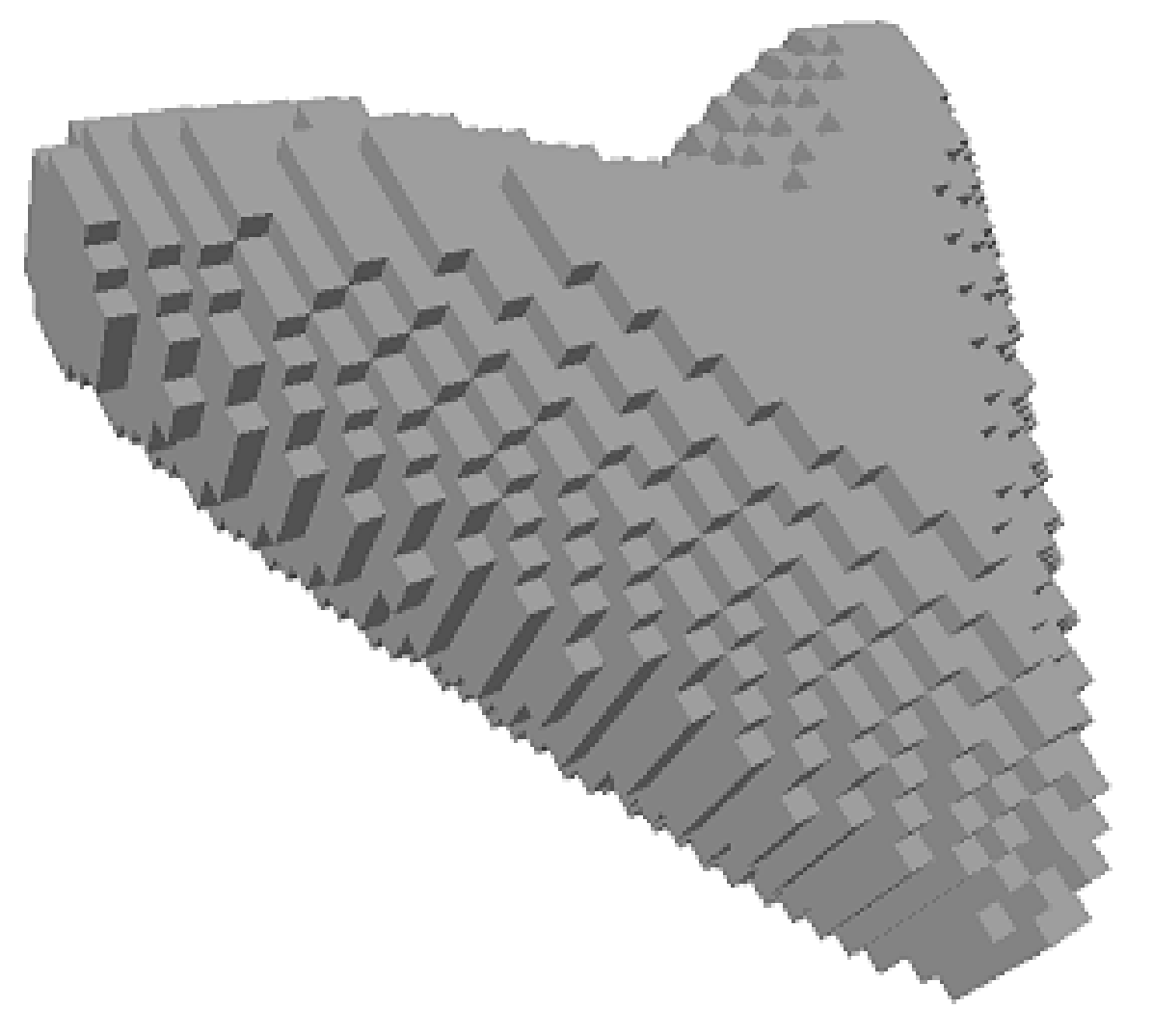}} 
	\subfloat[Target]{\label{fig:heart-1000}\includegraphics[width=\shapewidth]{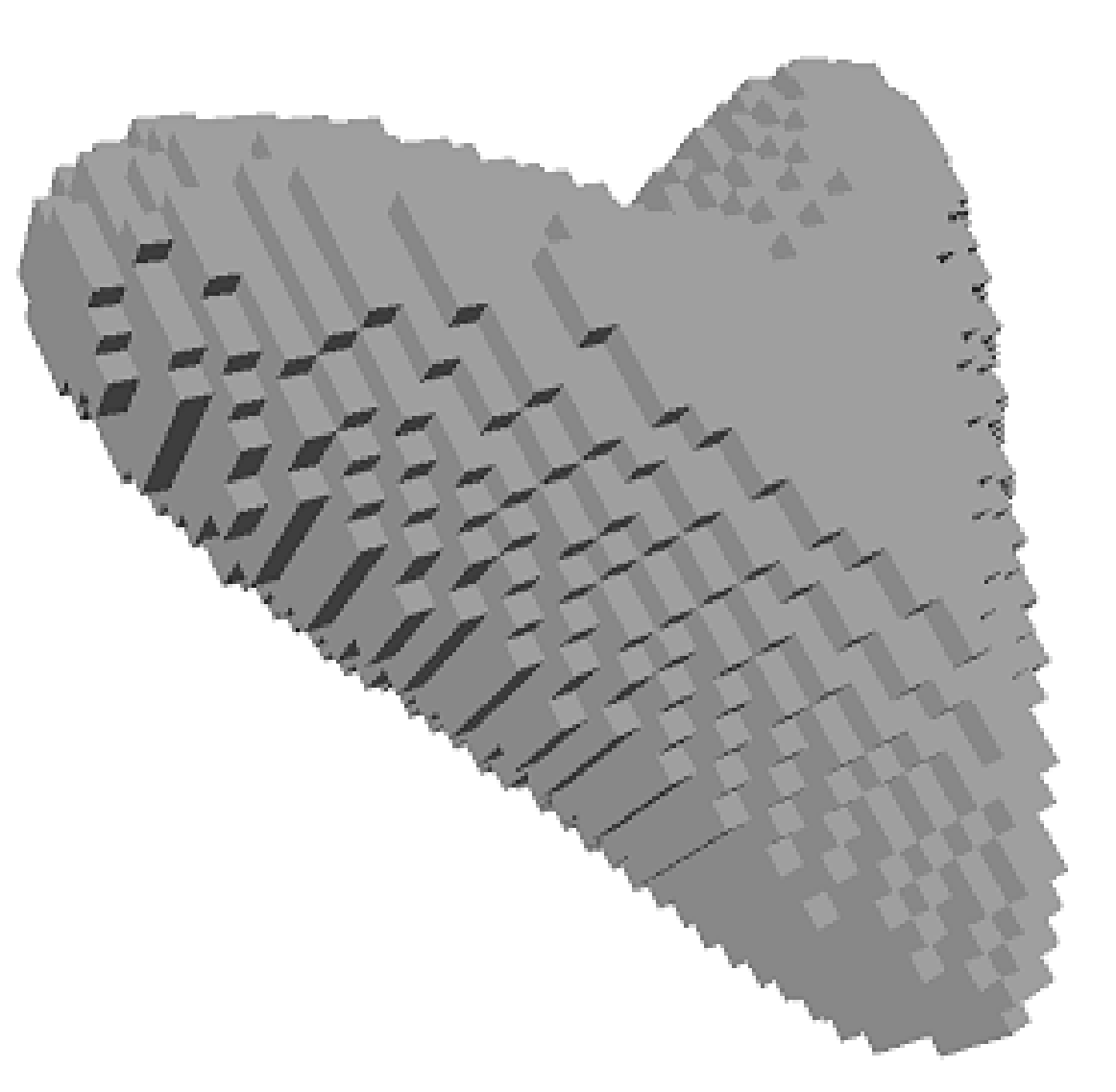}}
	\caption{Evolution of a 3D heart.}
	\label{fig:heart-evolution}
\end{figure}

Fig.~\ref{fig:cube-evolution} illustrates a sample of the evolved individuals from one cube experiment, Fig.~\ref{fig:star-evolution} similarly for the star experiment, and Fig.~\ref{fig:heart-evolution} for the heart experiment.  


\section{Rotation-based evolution}    

As previously mentioned, we have recently undertaken initial experimentation of EAs to design VAWT with a vector of integers representing the width of a turbine blade segment~\cite{PreenBull:2013,PreenBull:2014}. The fitness of each individual was scored as the maximum rotation speed achieved during the application of constant wind generated by a propeller fan after fabrication by a 3D printer. The rotation speed was measured in number of revolutions per minute (rpm) using a digital photo laser tachometer (PCE-DT62; PCE Instruments UK Ltd) by placing a $10\times2$~mm strip of reflecting tape on the individual. When measuring a single isolated VAWT, the individual was placed at 30~mm distance from the centre of a 30~W, 3500~rpm, 304.8~mm propeller fan generating 4.4~m/s wind speed.
                                                                                                                                                                         
\begin{figure}[t]
	\centering 
	\subfloat[{Basic superformula seed design. Genome: $m_1=6$, $n_{1,1}=5$, $n_{1,2}=30$, $n_{1,3}=10$, $m_2=4$, $n_{2,1}=10$, $n_{2,2}=10$, $n_{2,3}=10$}.]
	{%
		\includegraphics[width=1.0in]{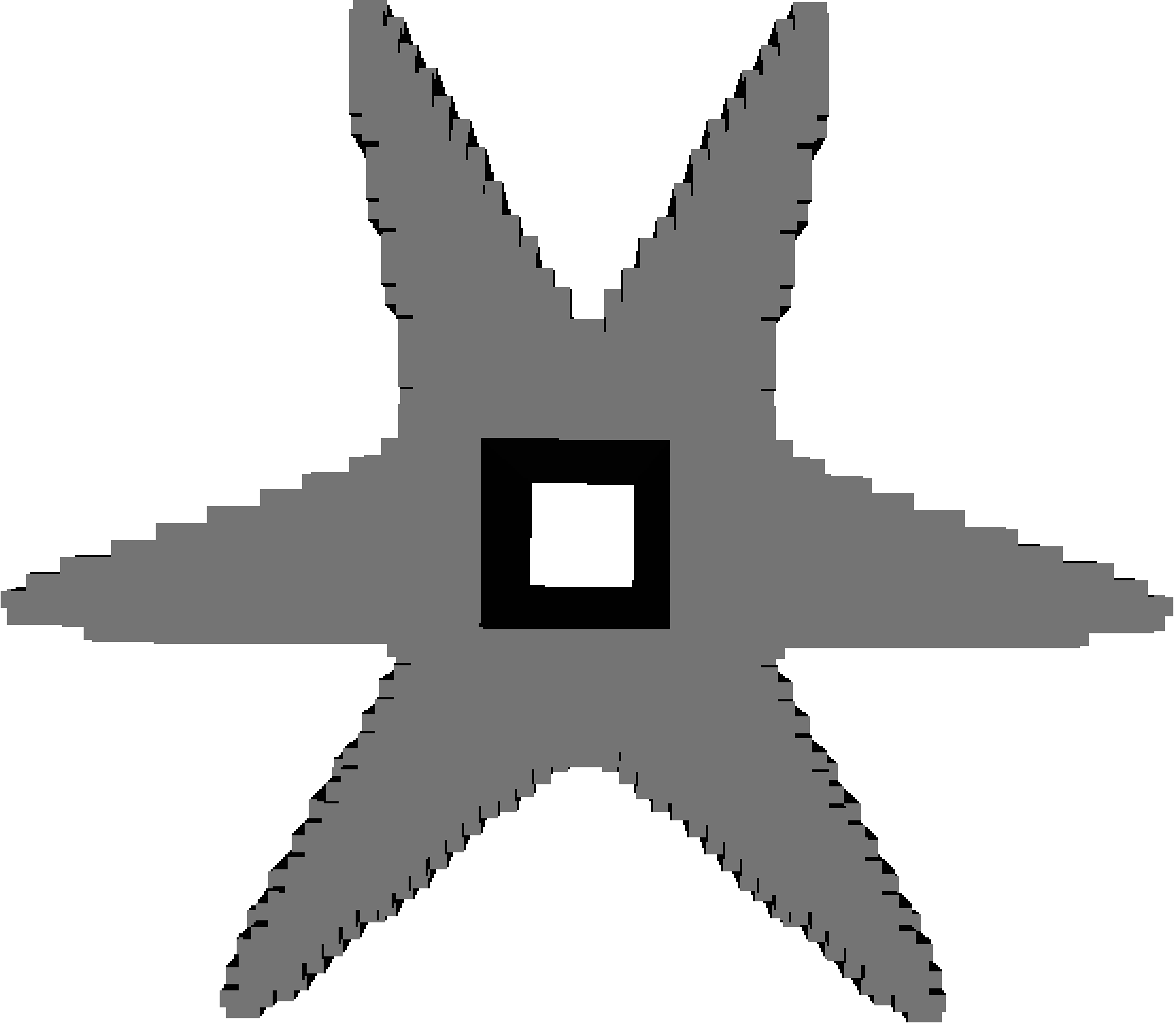} 
		\hspace{0.2in}
		\includegraphics[width=1.0in]{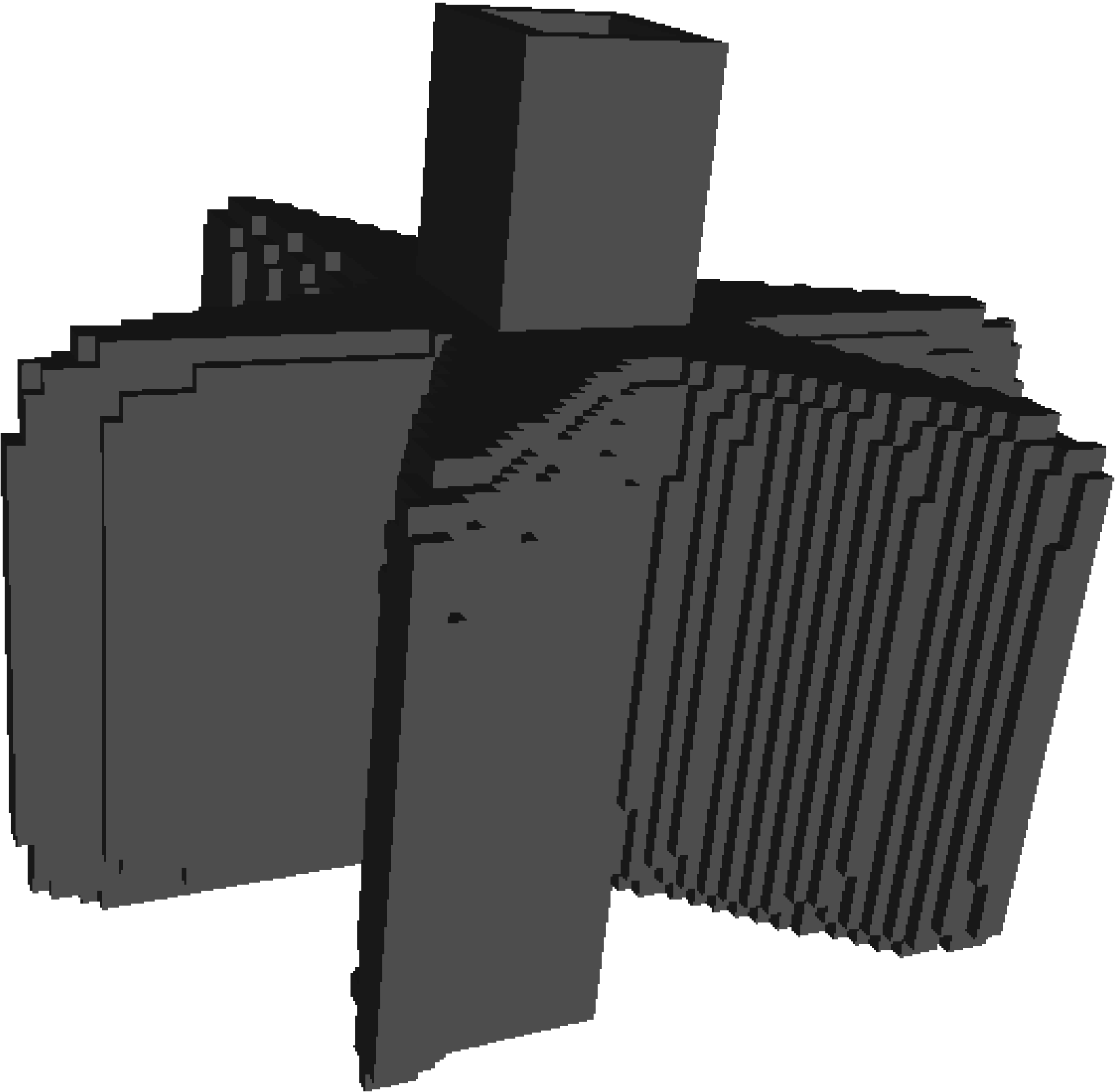}
		\label{fig:star-unsmoothed}
	}

	\subfloat[Basic superformula seed design with 3 Laplacian smoothing steps applied.]
	{%
		\includegraphics[width=1.0in]{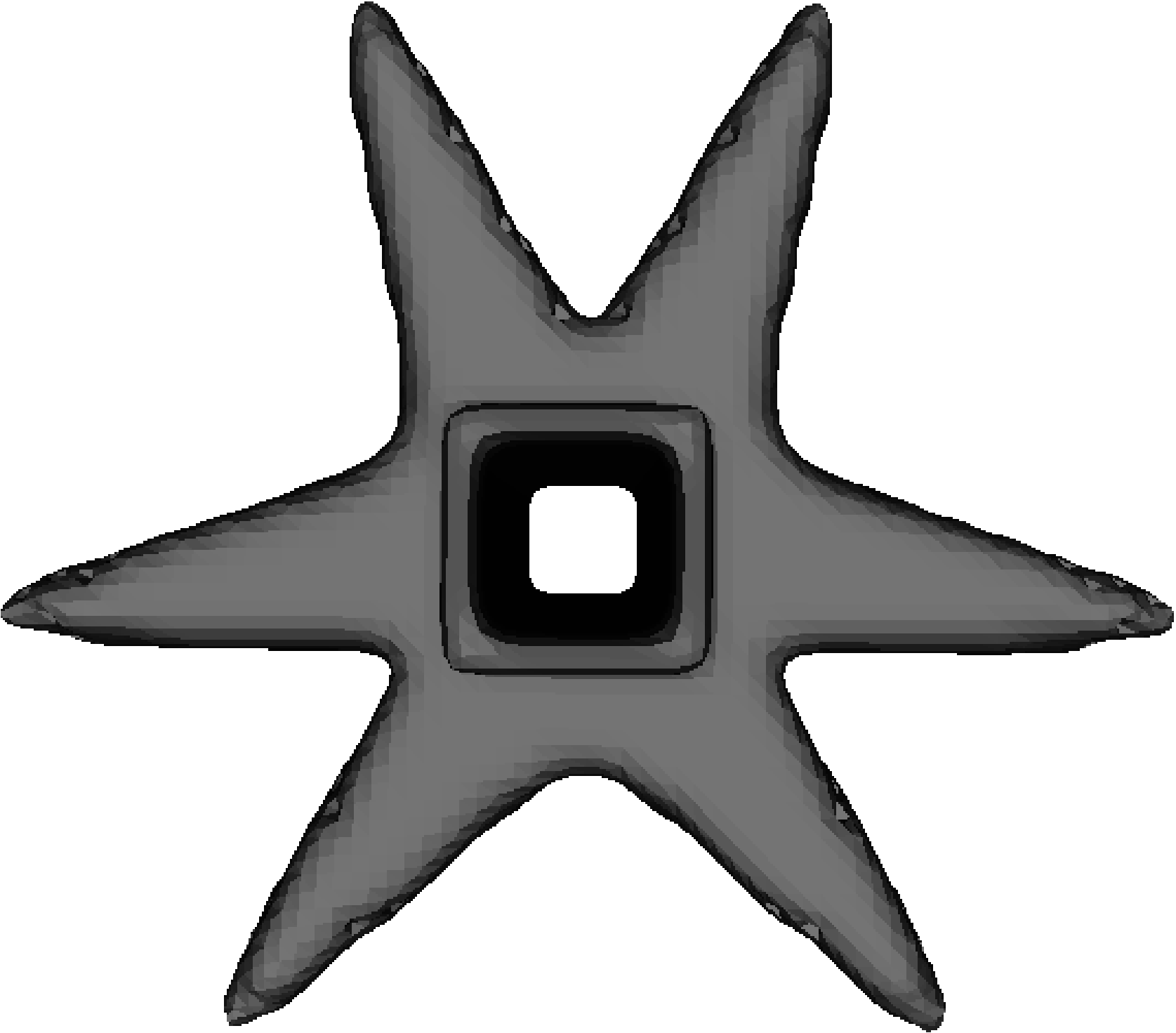} 
		\hspace{0.2in}
		\includegraphics[width=1.0in]{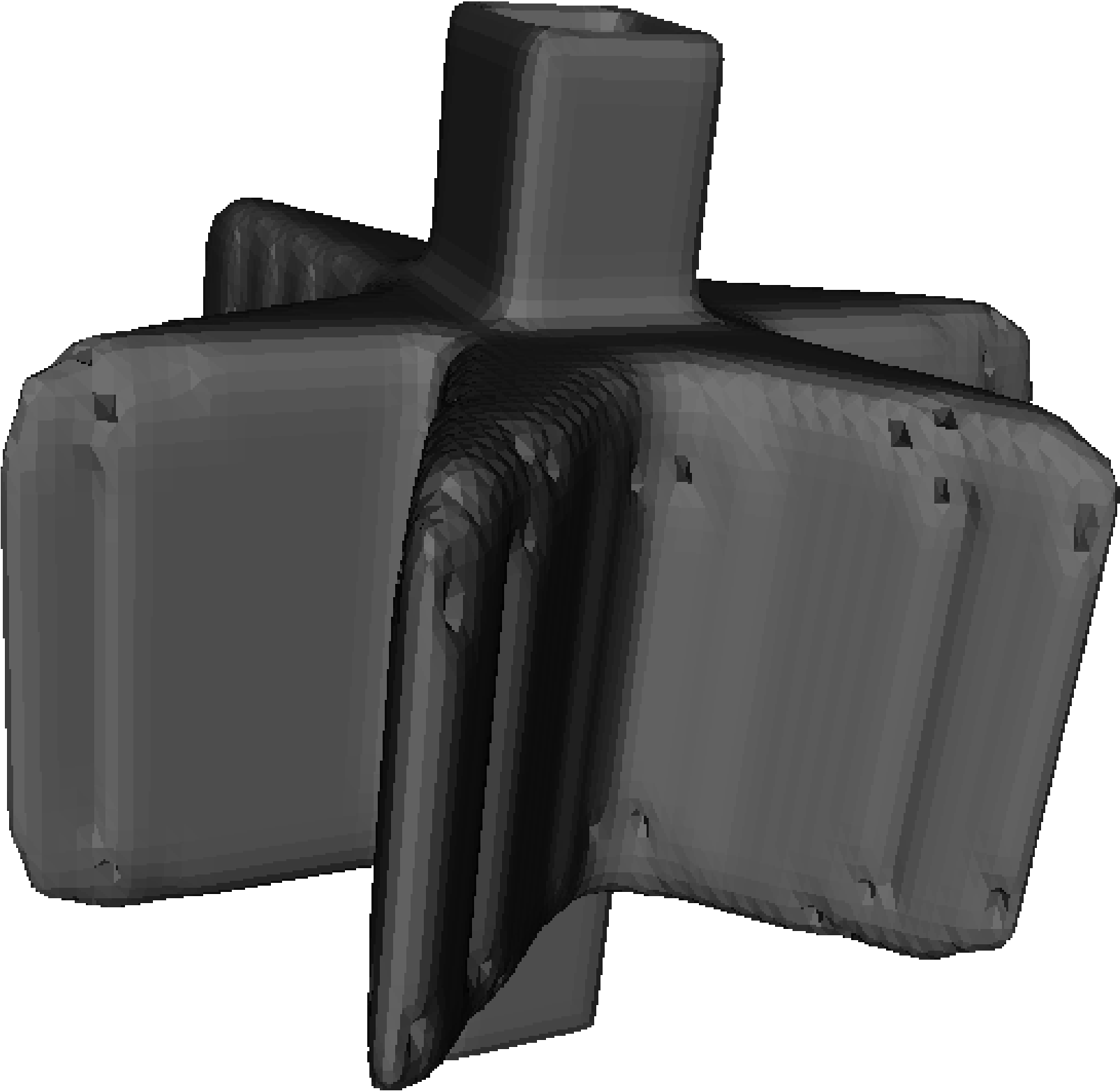}
		\label{fig:star-smoothed}
	}

	\subfloat[Basic superformula seed design smoothed and printed by a 3D printer; $50\times50\times50$~mm; 80~minutes printing time.]
	{%
		\includegraphics[width=1.4in]{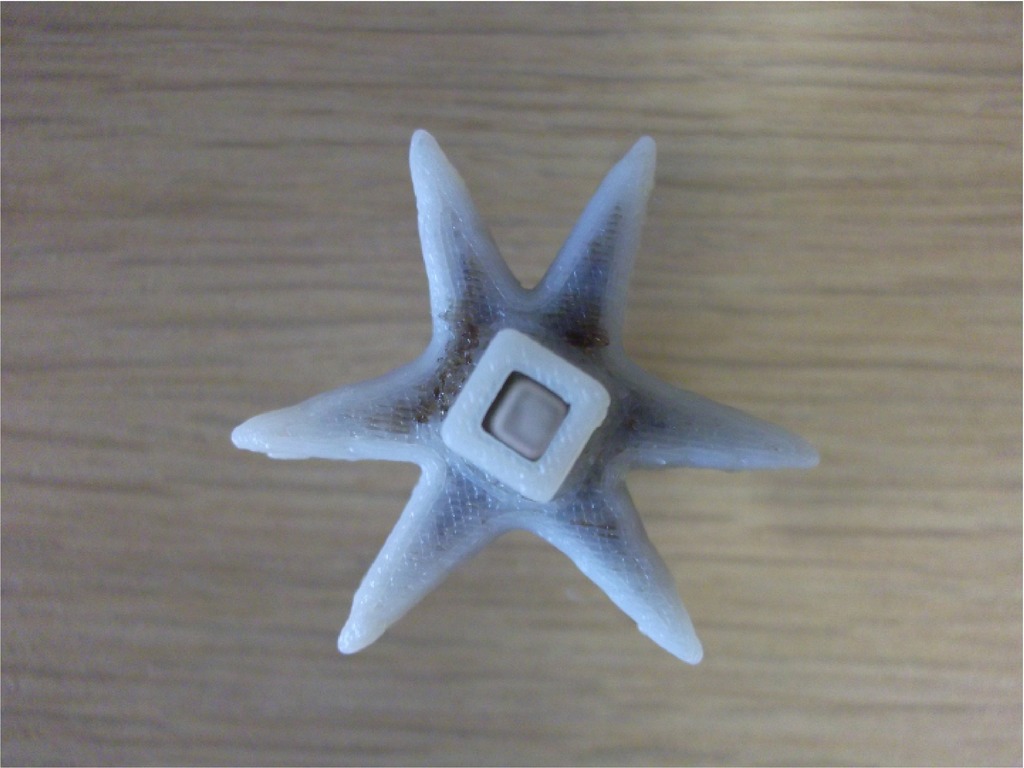} 
		\includegraphics[width=1.4in]{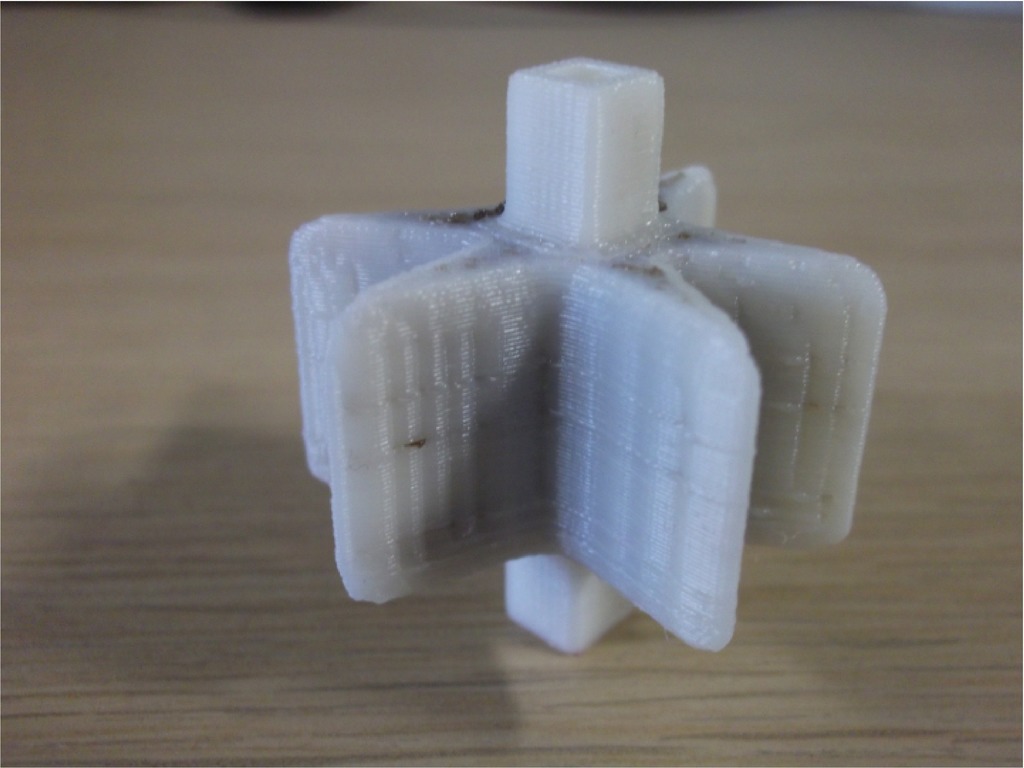}
		\label{fig:star-printed}
	}
	\caption{Basic superformula seed design.}
	\label{fig:star-vawt}
\end{figure}

One of the drawbacks of the representation used previously is that it assumes an underlying VAWT structure. In contrast, supershapes open the space of possible designs and yet can retain a compact encoding. As a first step towards the evolution of supershapes as VAWT, here a single supershape, as described previously, becomes a prototype VAWT. A workspace (maximum object size) of $50\times50\times70$~mm is used so that the instantiated prototype is small enough for timely production ($\sim80$~minutes) and with low material cost, yet large enough to be sufficient for fitness evaluation. The workspace has a resolution of $100\times100\times100$ voxels. A central platform is constructed for each individual to enable the object to be placed on to the evaluation equipment. The platform consists of a square torus, 2 voxels in width and with a centre of $10\times10$ empty voxels consistent through the $z$-axis, thus creating a hollow tube; see example in Fig.~\ref{fig:star-unsmoothed}.

When production is required, the 3D binary voxel array is converted to stereolithography (STL) format. Once encoded in STL, it then undergoes post-processing with the application of Laplacian smoothing using MeshLab\footnote{MeshLab is an open source, portable, and extensible system for the processing and editing of unstructured 3D triangular meshes.\ \url{http://meshlab.sourceforge.net}}; see example in Fig.~\ref{fig:star-smoothed}. Finally, the object is converted to printer-readable G-code and is subsequently fabricated by a Stratasys Dimension Elite printer using a polylactic acid ({PLA}) bioplastic. See example in Fig.~\ref{fig:star-printed}.

\begin{figure}[t]
	\centering 
	\includegraphics[width=\figwidth]{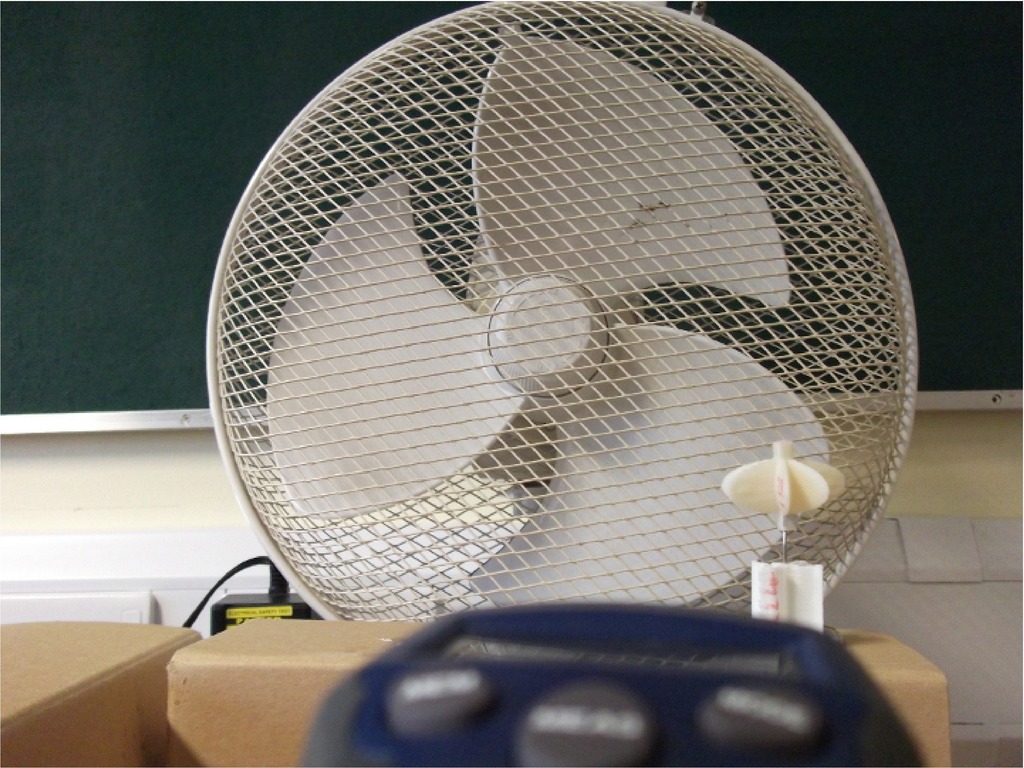}
	\caption{Basic superformula experimental setup; asymmetric airflow.}
	\label{fig:setup-single}
\end{figure}

The fitness computation for each individual is the maximum rotation speed achieved during the application of constant wind generated by a propeller fan after fabrication by a 3D printer similar to the previous experiments as described above. Two experiments are conducted; the first using the set of 8 basic superformula parameters, and the second with an extended set of 16. In the basic superformula experiment, the initial population consists of the star individual from Fig.~\ref{fig:star-vawt} and 19 other individuals whose parameters are each those of the star mutated by a random number in the range [-5,5]; that is, $P=20$ and each individual consists of 8 superformula parameters. Due to the large degree of symmetry with the basic superformula, each candidate VAWT is positioned with an asymmetric air flow of 4.4~m/s; see experimental configuration in Fig.~\ref{fig:setup-single}, which shows the propeller fan and the individual placed at 30~mm distance and offset by 100~mm from the centre.
 
All initial individuals are subsequently fabricated and evaluated. Thereafter, a generational GA forms the next generation using the evolutionary operators as described for the target-based experiment and with a limited form of elitism, promoting the single fittest individual to the next generation. 

\begin{figure}[t]
	\centering 
	\subfloat[Top view]{\includegraphics[width=1.6in]{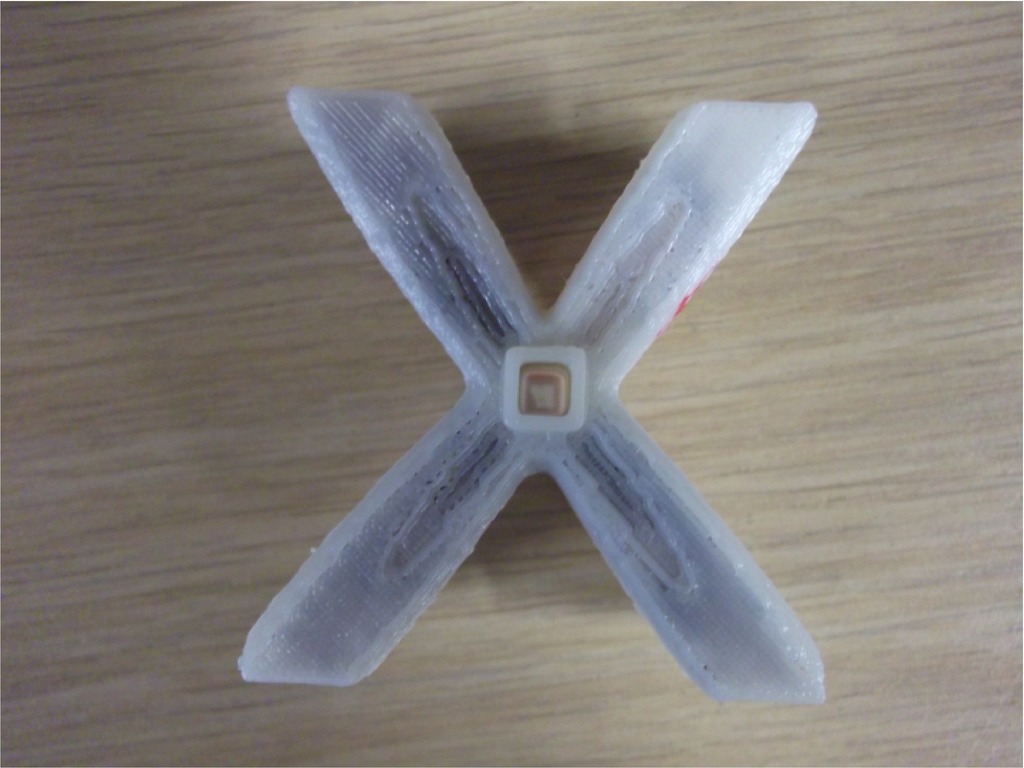}}
	\subfloat[Side view]{\includegraphics[width=1.6in]{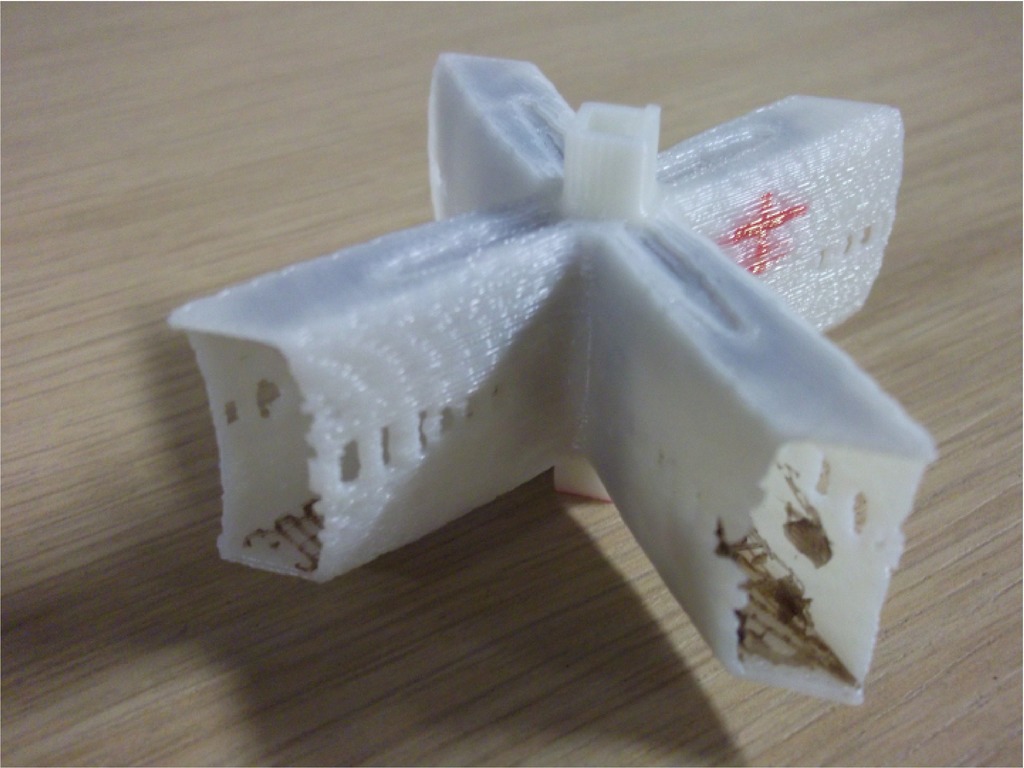}}
	\caption{Basic superformula fittest individual after 4 generations. 537~rpm.}
	\label{fig:ga-prevbest}

	\subfloat[Top view]{\includegraphics[width=1.6in]{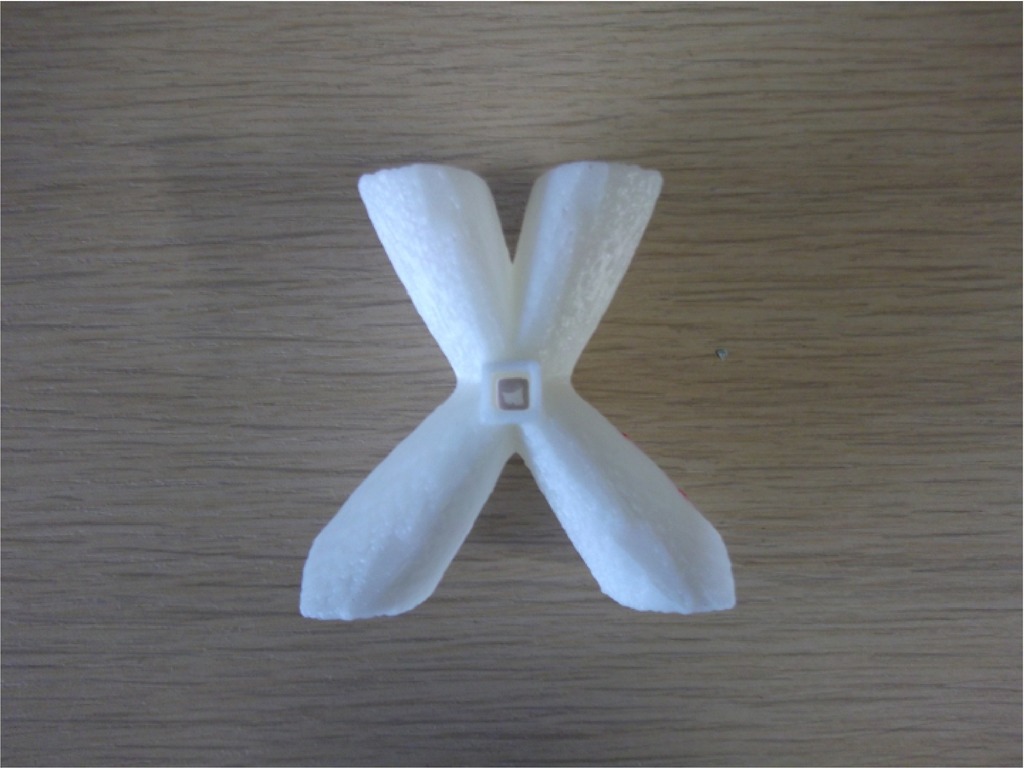}}
	\subfloat[Side view]{\includegraphics[width=1.6in]{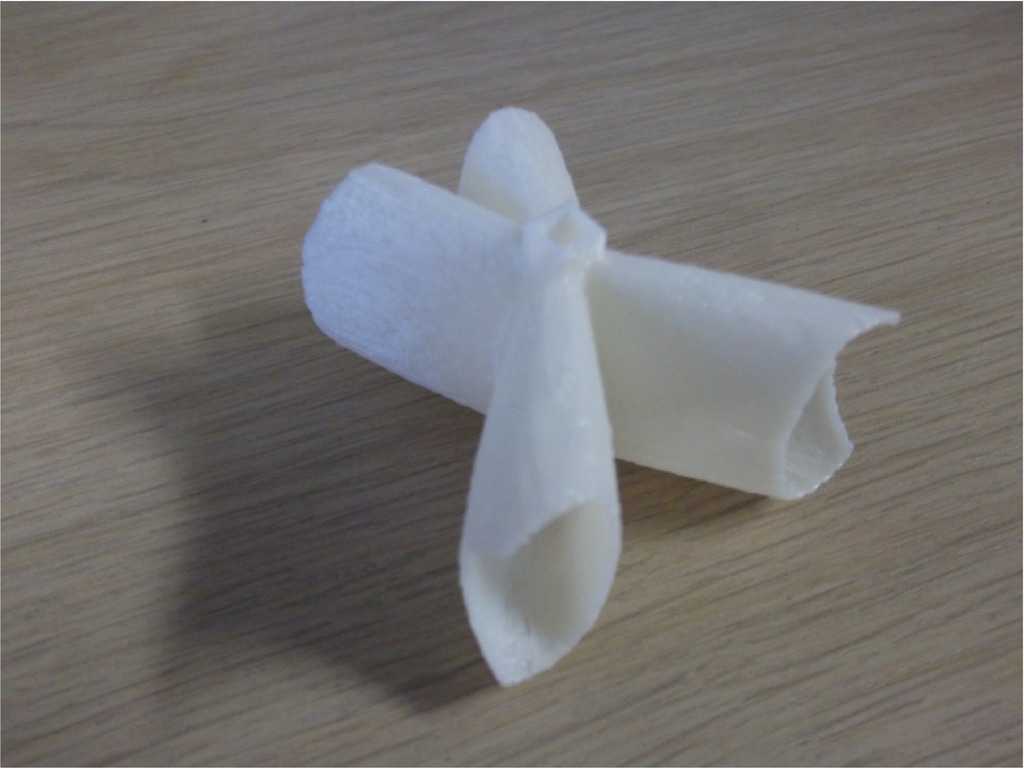}}
	\caption{Basic superformula fittest individual after 5 generations. 581~rpm.}
	\label{fig:ga-best}
\end{figure}

The fittest evolved individual after 4 generations from the basic superformula experiment is shown in Fig.~\ref{fig:ga-prevbest}. The parameters to the superformula specify the length of the blades in addition to the frequency and the population has evolved an individual that forms an `X' shape where the blades extend beyond the length of the workspace. As the blades extend beyond the workspace they are no longer drawn/fabricated and so the hollowness of the shape can be observed. It appears that evolution has identified that longer blades are more efficient under the current experimental conditions and this is also observed with an increase in the average length of the blades throughout the population. Furthermore, the reduction in number of blades from the initial 6 to 4 indicates that fewer blades may be more efficient. In Fig.~\ref{fig:ga-best} the fittest evolved individual after 5 generations is shown. As can be seen, overall the shape is more rounded and two of the blades from the `X' have merged closer together in a step towards a 3 bladed `Y' shape, resulting in a lighter weight design with an increase in rotation speed.

\begin{figure}[t]
	\centering 
	\subfloat[{Extended superformula seed design. Genome: $m_1=0$, $n_{1,1}=0$, $n_{1,2}=0$, $n_{1,3}=1$, $m_2=0$, $n_{2,1}=0$, $n_{2,2}=0$, $n_{2,3}=1$, $t_1=0$, $t_2=6$, $d_1=0.5$, $d_2=0.7$, $c_1=4$, $c_2=1$, $c_3=0.4$, $r_0=50$.}]
	{%
		\raisebox{0.5\height}{\includegraphics[width=1.0in]{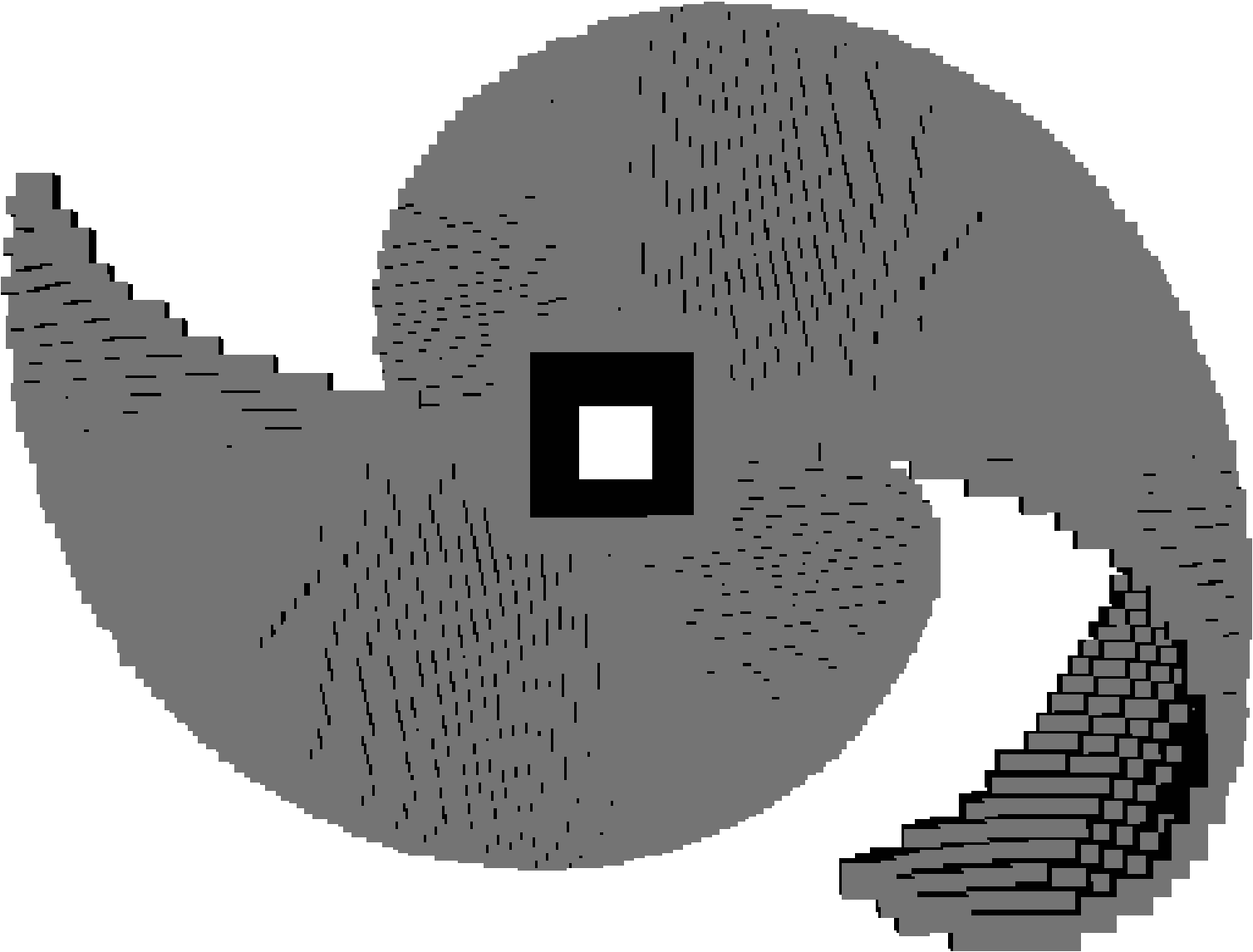}}
		\hspace{0.2in}
		\includegraphics[width=1.0in]{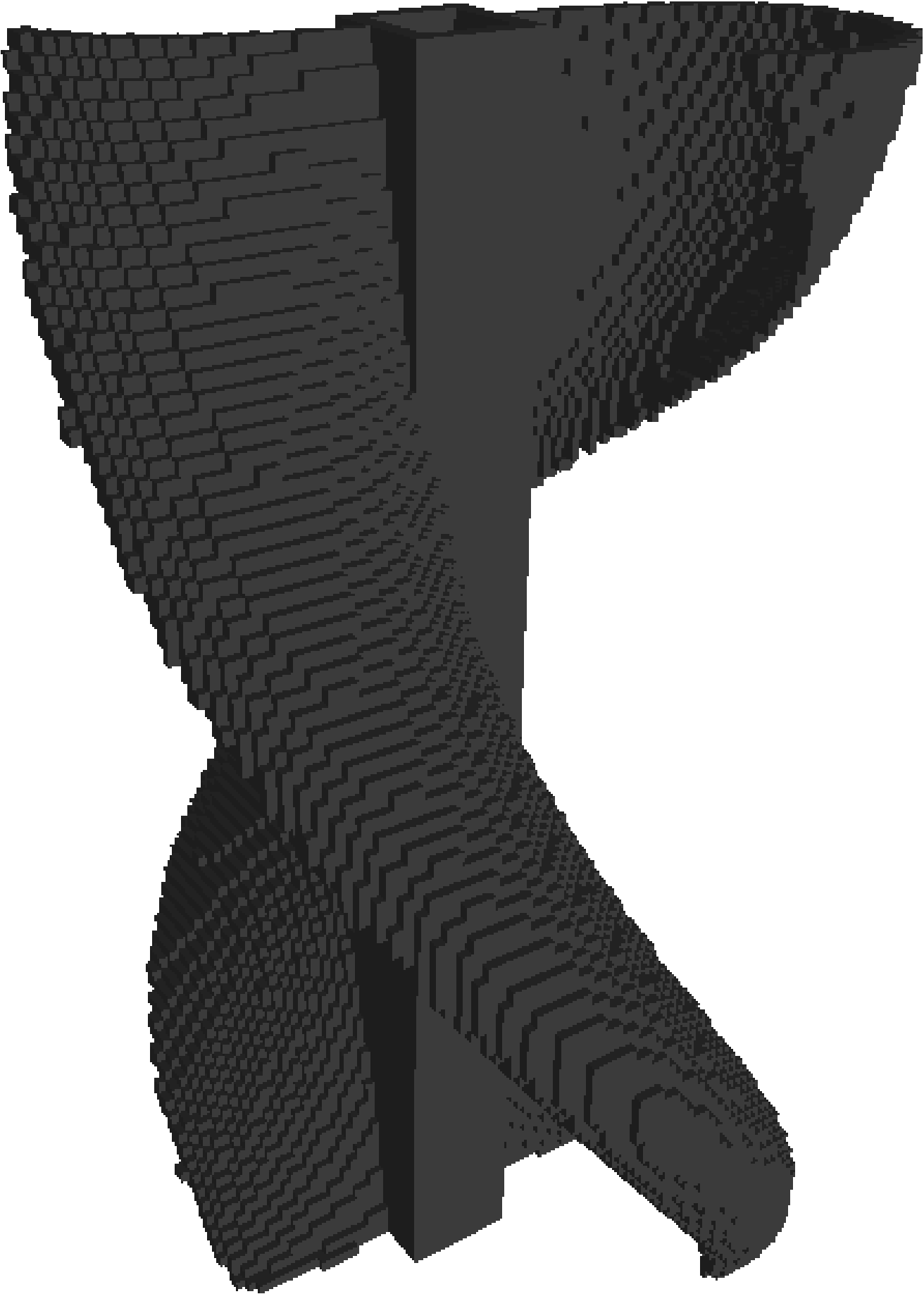}
	}

	\subfloat[Extended superformula seed design with 50 Laplacian smoothing steps applied.]
	{%
		\raisebox{0.5\height}{\includegraphics[width=1.0in]{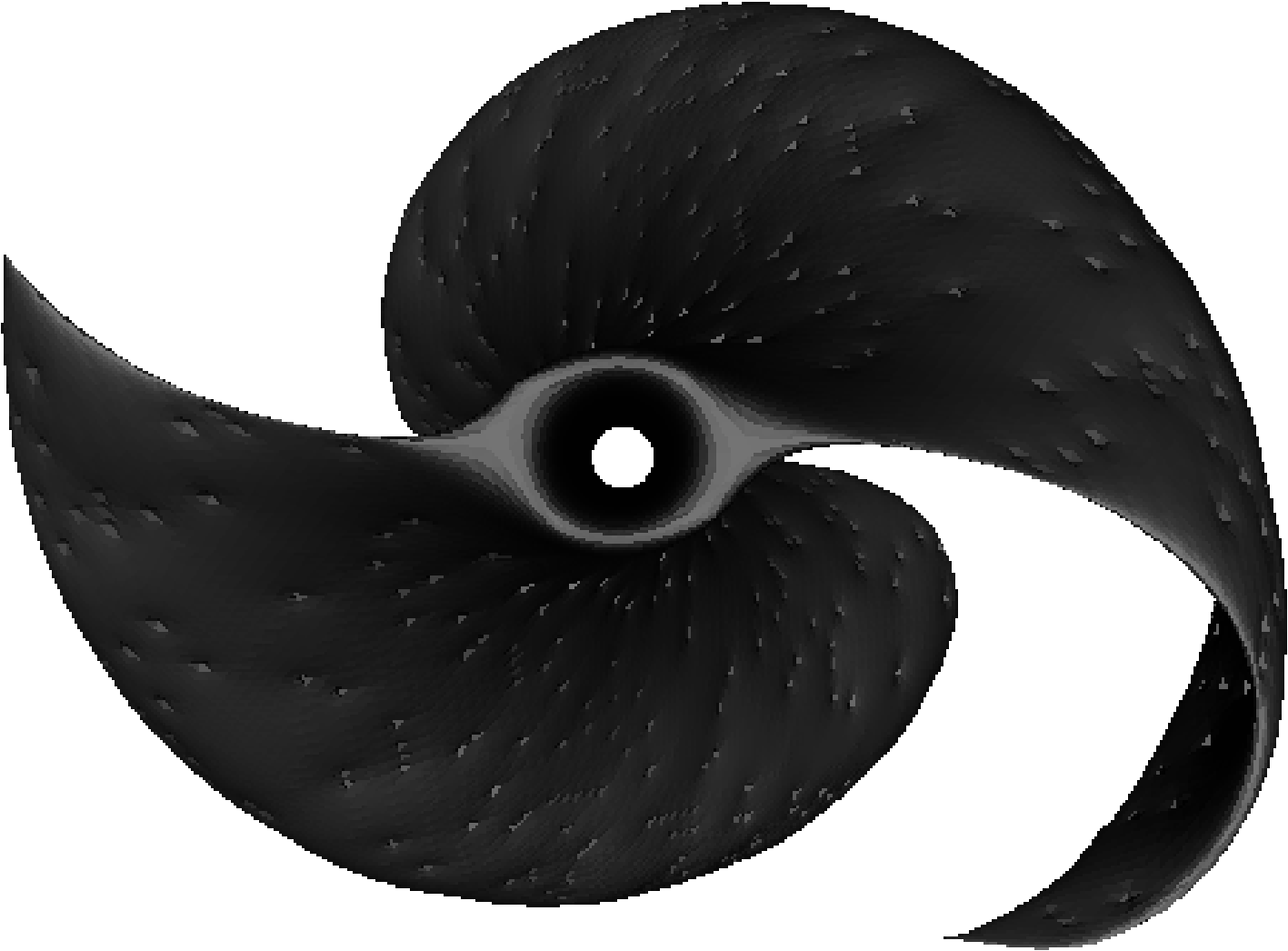}}
		\hspace{0.2in}
		\includegraphics[width=1.0in]{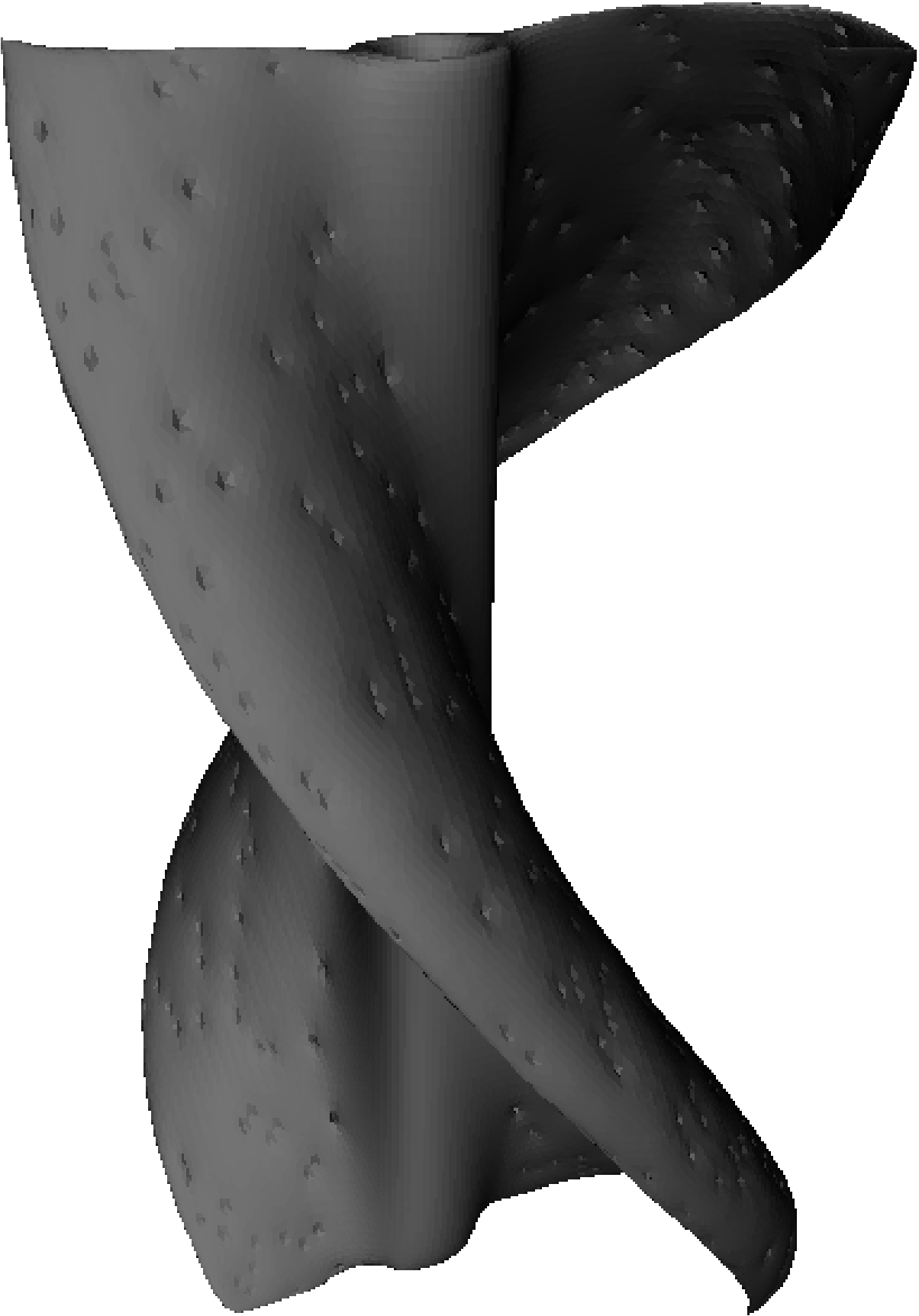}
	}

	\caption{Extended superformula seed design.}
	\label{fig:twist-seed}
\end{figure}

\begin{figure}[t]
	\centering 
	\includegraphics[width=1.6in]{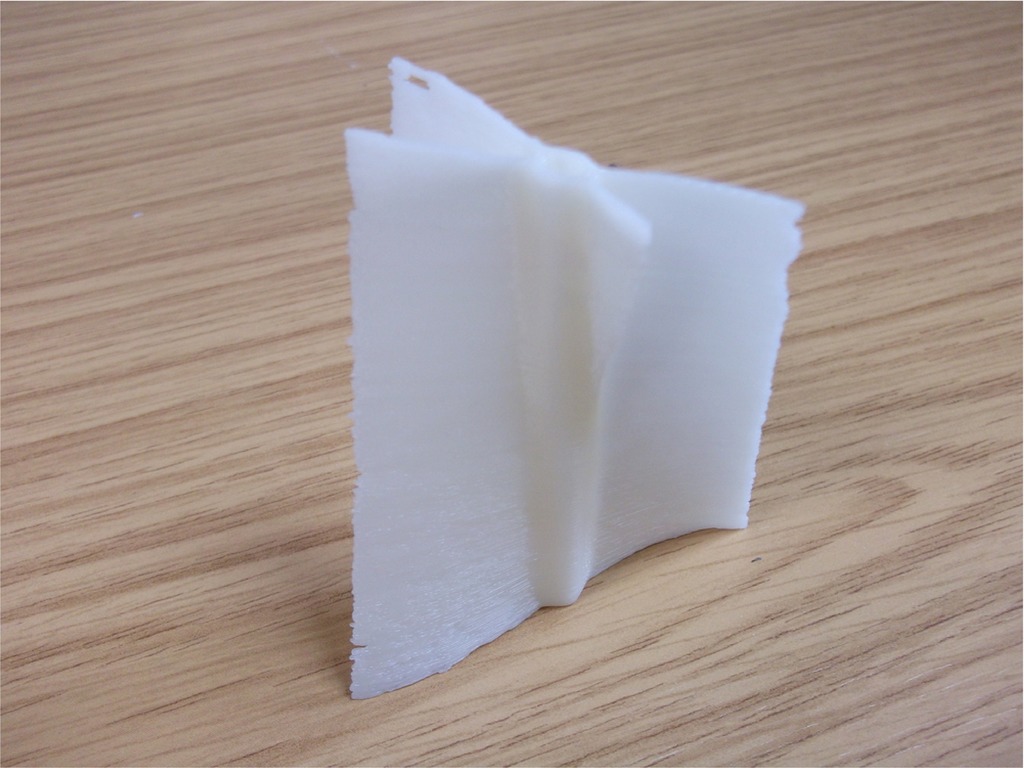}
	\includegraphics[width=1.6in]{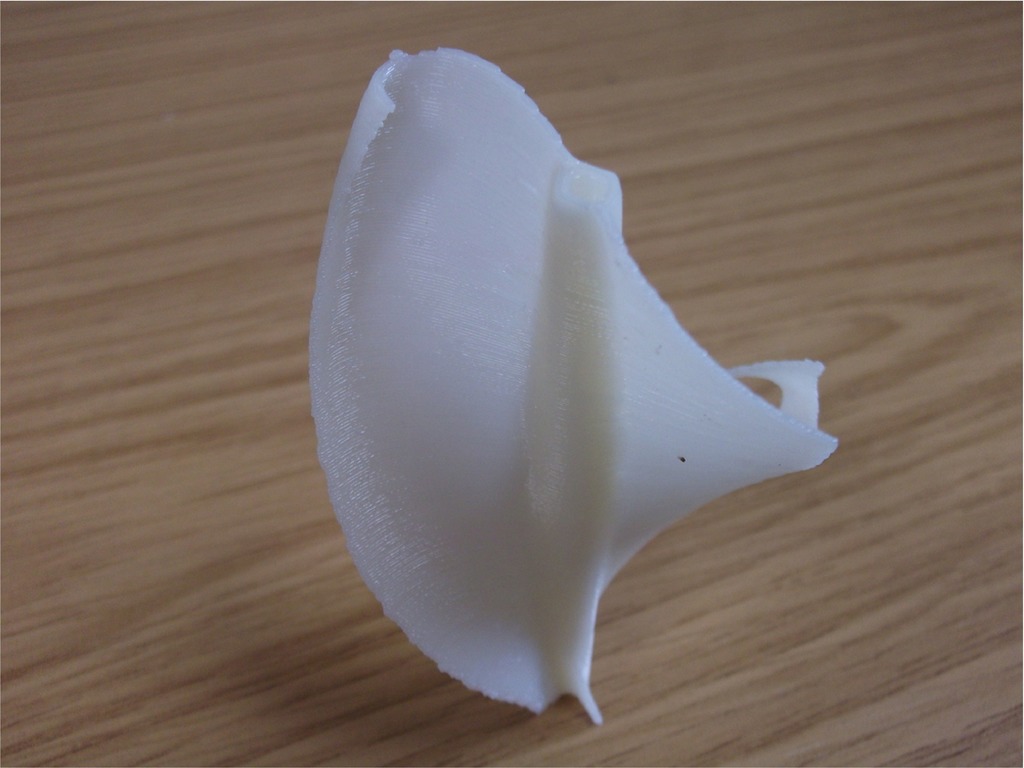} \\
	\includegraphics[width=1.6in]{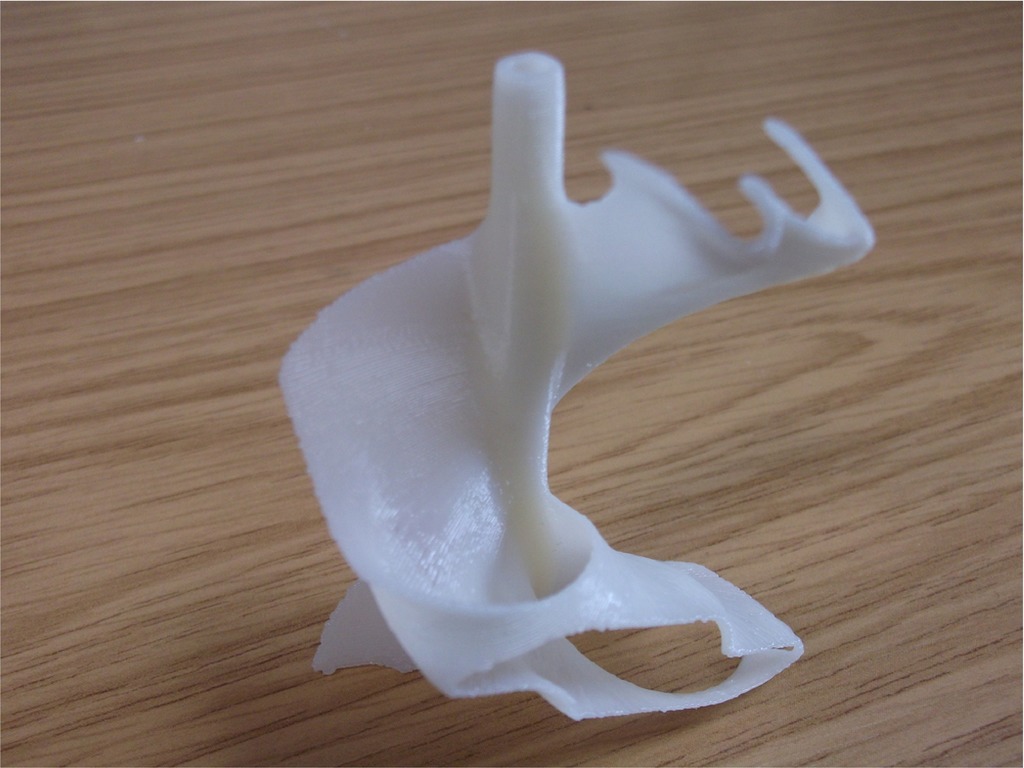}
	\includegraphics[width=1.6in]{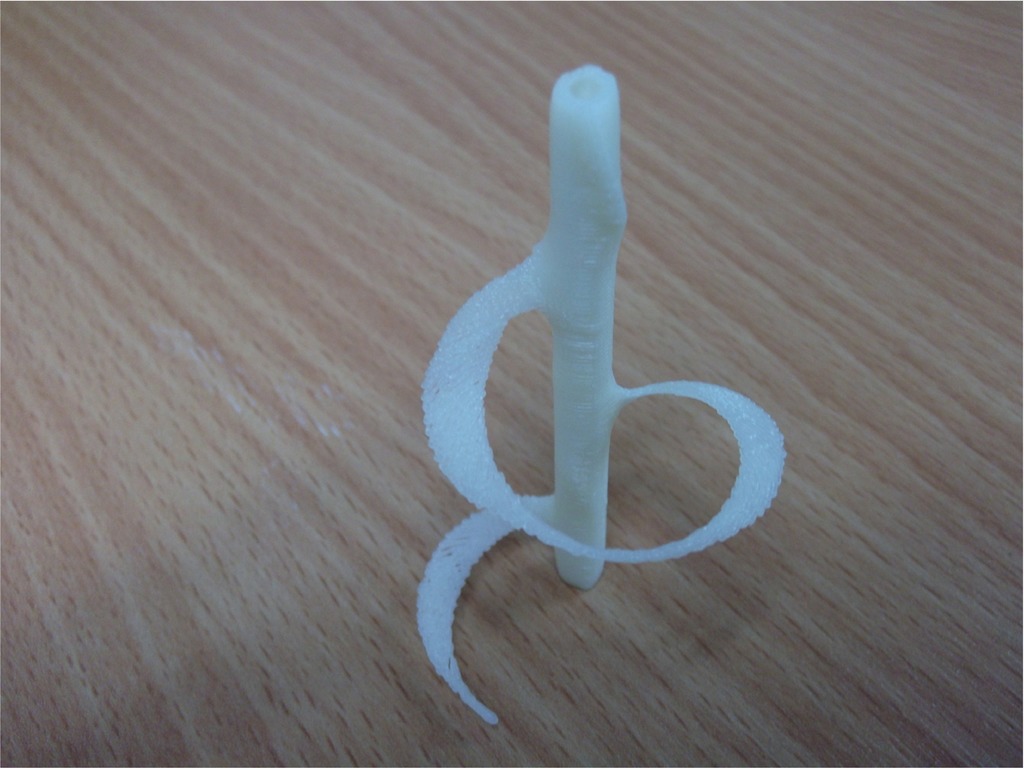}
	\caption{Example individuals from the initial population of the extended superformula experiment.}
	\label{fig:twist-initial}
\end{figure}

\begin{figure}[t]
	\centering 
	\includegraphics[width=\figwidth]{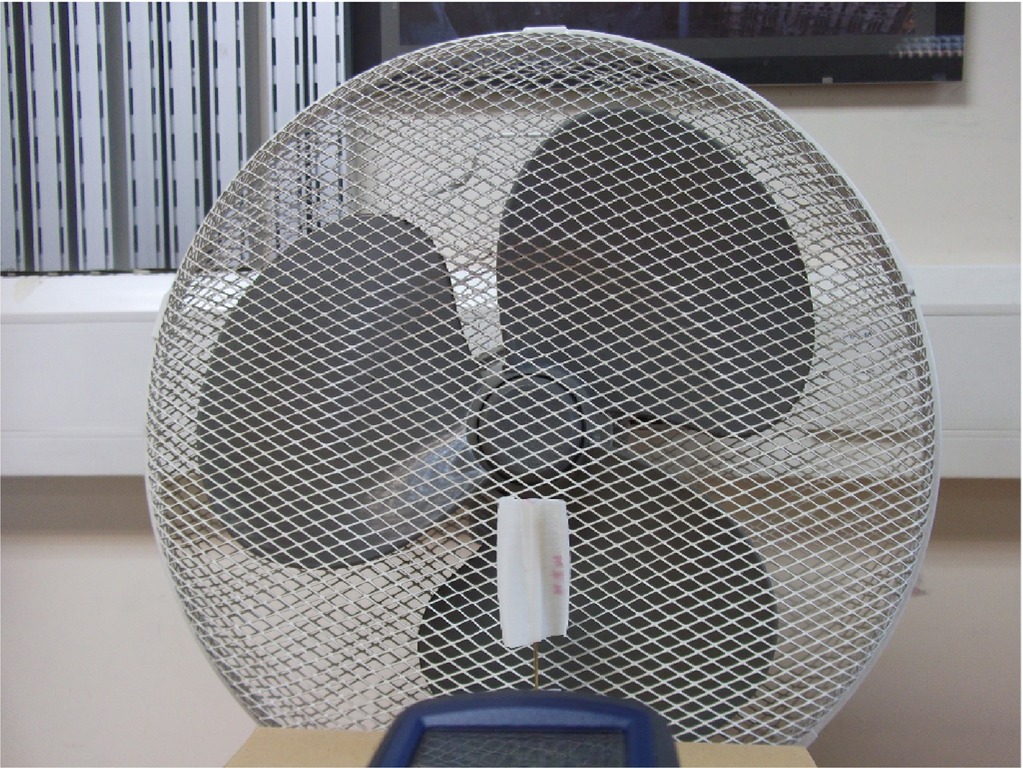}
	\caption{Extended superformula experimental setup; symmetric airflow.}
	\label{fig:setup-twist}
\end{figure}

In the extended superformula experiment, the degrees of freedom are increased by 7 additional parameters that enable more complex supershapes, such as the m\"obius strip in Fig.~\ref{fig:mobius}. A single further parameter is used to control the size of the supershape, $r_{0}$, for a total of 16. In this experiment the initial generation is seeded with designs mutated around parameters that approximate an S-shaped VAWT and all designs undergo 50 Laplacian smoothing steps before being fabricated (see seed design in Fig.~\ref{fig:twist-seed}.) To form each initial individual, the original 8 parameters from the seed design are mutated by a random number in the range [-10,10] and the additional 8 parameters mutated by a random number in the range [-1,1]; any supershape produced with fewer than 1\% active voxels is discarded and another generated. This creates a range of very diverse shapes; see examples from the initial population in Fig.~\ref{fig:twist-initial}. As such deformations enable asymmetrical designs, each individual is placed in a more natural airflow, centre of the propeller fan, see Fig.~\ref{fig:setup-twist}. The GA proceeds as before, however when mutation occurs for one of the additional 8 parameters, the value is altered by a random number in the range [-0.5,0.5], rather than the usual [-5,5], due to their sensitivity.
  
\begin{figure}[t]
	\centering 
	\subfloat[Top view]{\includegraphics[width=1.6in]{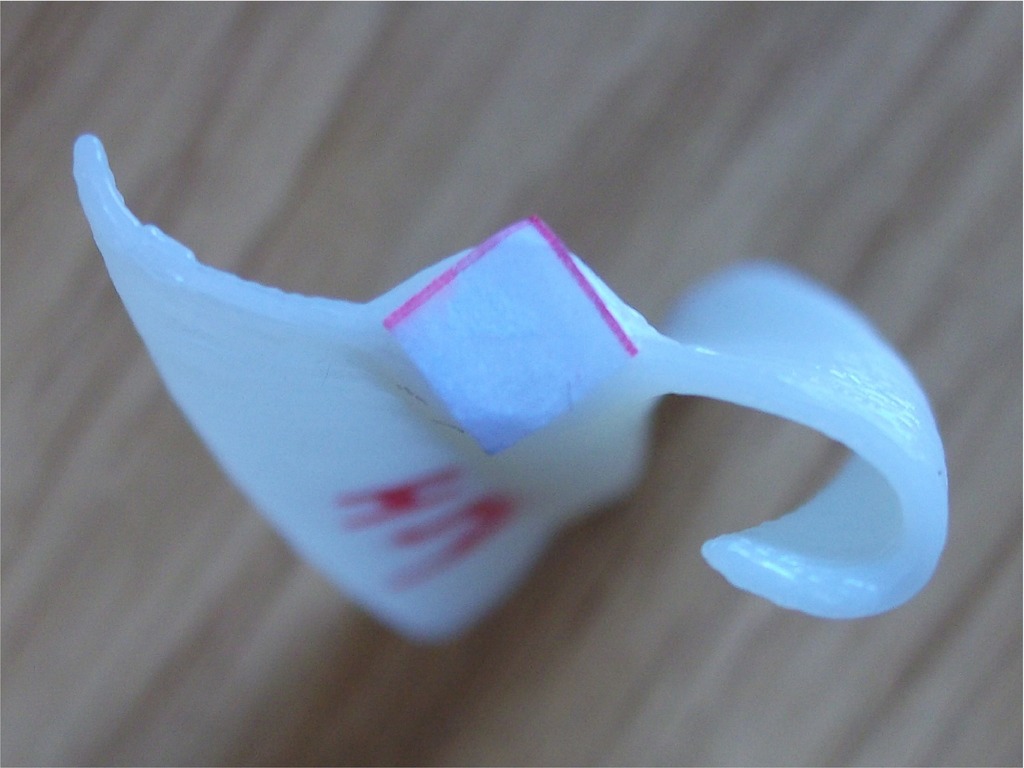}}
	\subfloat[Side view]{\includegraphics[width=1.6in]{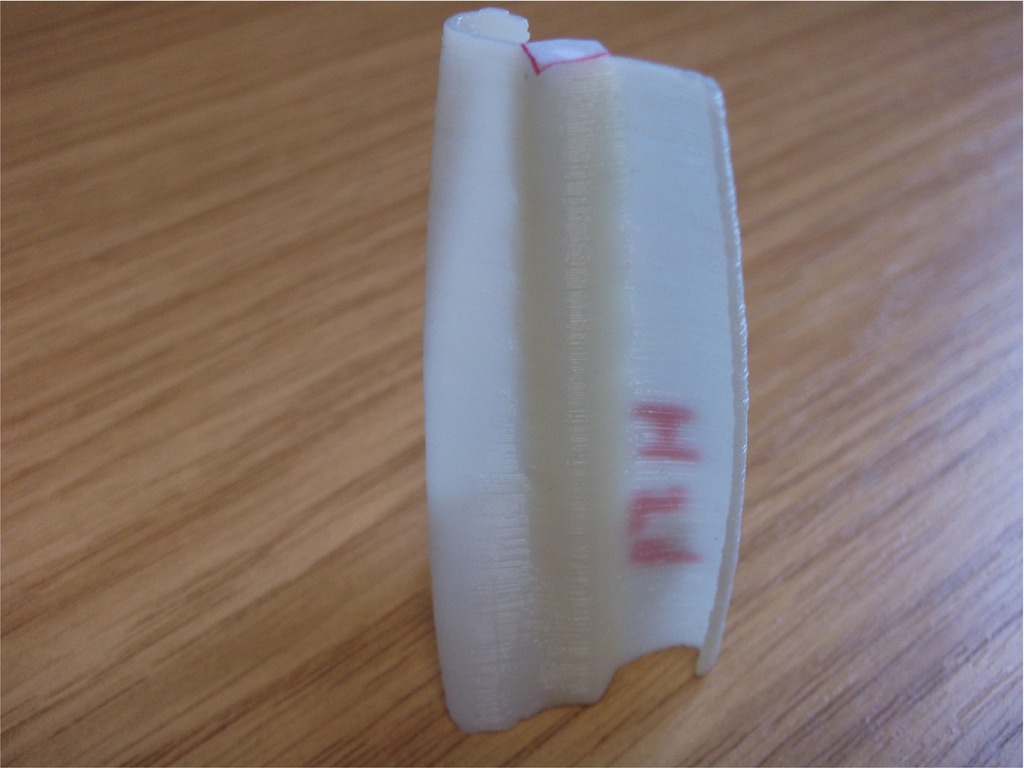}}
	\caption{Extended superformula fittest individual after 3 generations. 986~rpm. Dimensions: 66~mm height; 28~mm diameter; 1~mm thickness.}
	\label{fig:twist-g3}
\end{figure}

\begin{figure}[t]
	\centering 
	\subfloat[Top view]{\includegraphics[width=1.6in]{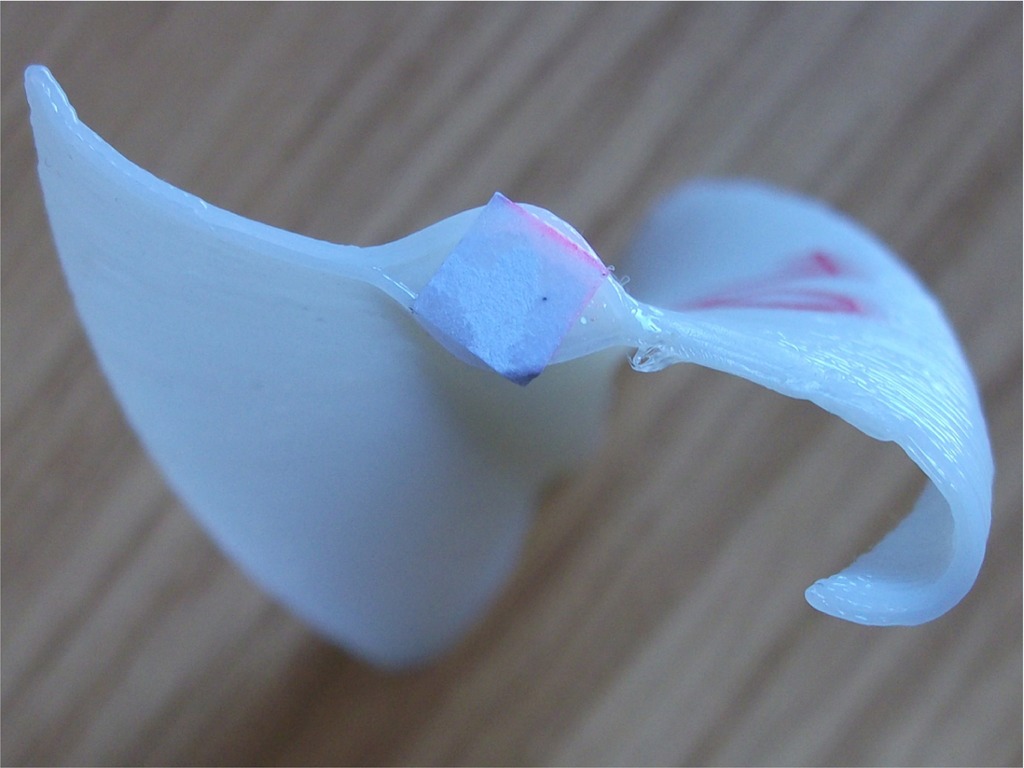}}
	\subfloat[Side view]{\includegraphics[width=1.6in]{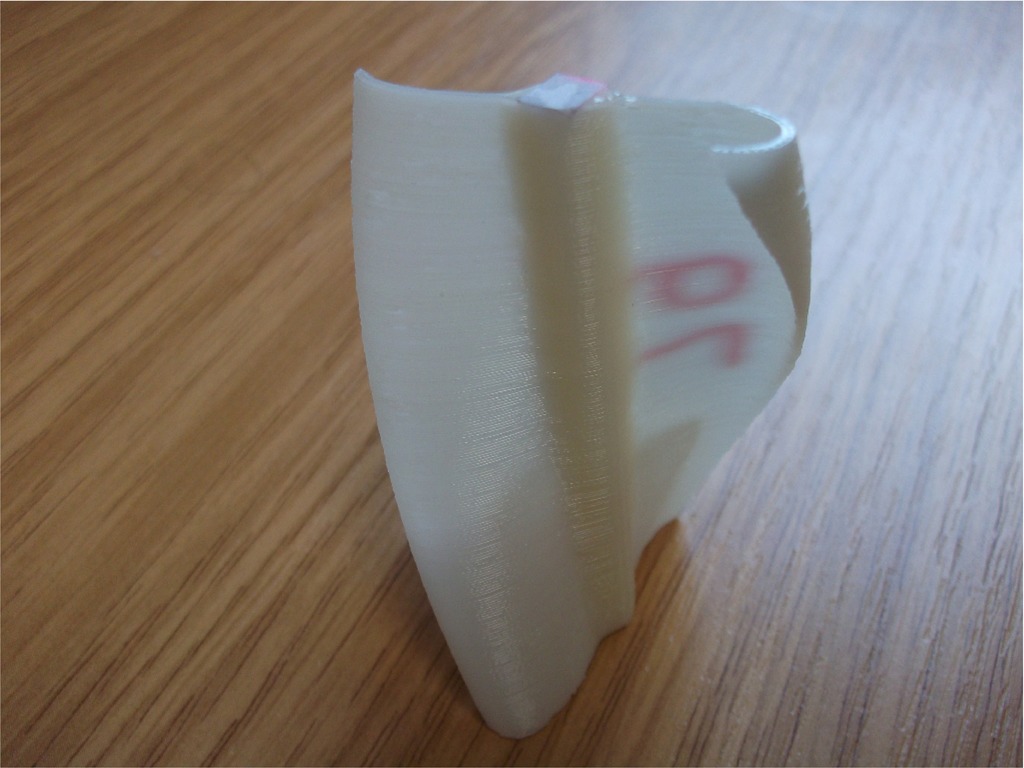}}
	\caption{Extended superformula fittest individual from 4th/5th generation. 1070~rpm. Dimensions: 66~mm height; 39~mm diameter; 1~mm thickness.}
	\label{fig:twist-g4}
\end{figure}

The fittest individual from the first generation had a fitness of 931~rpm; this increased to 985~rpm in the second generation; 986~rpm after three generations (see design in Fig.~\ref{fig:twist-g3}); and 1070~rpm after the fourth and fifth generations (see design in Fig.~\ref{fig:twist-g4}), showing that artificial evolution is capable of iteratively increasing the aerodynamic efficiency (i.e., rotational speed) of instantiated VAWT prototypes represented as supershapes.

\section{Conclusions and future work}

This paper has shown that it is possible to evolve a vector of reals that are used as superformula parameters to generate 3D objects. Target-based evolution was used to explore the ability of Gielis superformula to create complex objects. The experiments showed that with target-based evolution very closely matching designs can be identified. In addition, a methodology for the physical evolution of supershapes as VAWT has been introduced. A significant advantage of the approach over alternative representations is that it makes no assumptions about the underlying VAWT structure whilst maintaining the simplicity and compactness of the encoding, which may be amenable for future use in a surrogate-assisted approach.

The use of 3D printing to physically instantiate candidate designs completely avoids the use of 3D computer simulations, with their associated processing costs and modelling assumptions. In this case, 3D CFD analysis was avoided, but the approach is equally applicable to other real-world optimisation problems, for example, those requiring computational structural dynamics or computational electromagnetics simulations. We anticipate that in the future such `design mining' approaches will yield unusual yet highly efficient designs that exploit characteristics of the environment and/or materials that are difficult to capture formally or in simulation. This has the potential to place knowledge discovery at the core of engineering design, particularly within an iterative framework such as in agile approaches.

The fabrication of a candidate design with existing 3D printing technology still requires a considerable amount of time however ($\sim90$~minutes for the small designs here.) Therefore, techniques to speed up the process similar to those used in expensive numerical simulations are important. For example, multiple 3D printers can easily be used to perform parallel fabrication. In addition, although here the print resolution used to build the prototypes was set at the printer default, the resolution can be adjusted to provide coarser designs at a faster rate for preliminary studies (e.g., for early evolutionary candidates), or slower higher resolution prints for more subtle optimisation. Thus 3D printing offers a range of ways to customise the evolutionary instantiation to the design task. Algorithmic improvements to reduce the convergence time remain an important area of future research, in particular the use of surrogate models to reduce the number of fabrications.

Future work will include the use of the power generated by the VAWT prototypes as the fitness computation under various wind tunnel conditions; the coevolution of arrays, including turbine positioning; the application of surrogate modelling to reduce the number of fabrications; examination of the effect of seeding the EA with a given design; investigation of alternative 3D representations and the production of 1:1 scale designs.

The issue of scalability also remains an important future area of research. When increasing the scale of designs it is widely known that the changes in dimensionality will greatly affect performance, however it remains to be seen how performance will change in the presence of other significant factors such as turbine wake interactions in the case of arrays. One solution is to simply use larger 3D printing and wind-tunnel capabilities whereby larger designs could be produced by the same method. On the opposite end of the spectrum, micro-wind turbines that are 2~mm in diameter or smaller can be used to generate power, e.g., for wireless sensors~\citep{Howey:2011}, and in this case more precise 3D printers would be required. Moreover, wind turbines can find useful applications on any scale, e.g.,\ a recent feasibility study~\citep{Park:2012} for powering wireless sensors on cable-stayed bridges examined turbines with a rotor diameter of 138~mm in wind conditions with an average of 4.4~m/s (similar to the artificial wind conditions used in this paper.)

If the recent speed and material advances in rapid-prototyping continues, along with the current advancement of evolutionary design, it will soon be feasible to perform a wide-array of automated complex engineering optimisation {\it in situ}, whether on the micro-scale (e.g.,\ drug design), or the macro-scale (e.g.,\ wind turbine design). That is, instead of using mass manufactured designs, EAs will be used to identify bespoke solutions that are manufactured to compensate and exploit the specific characteristics of the environment in which they are deployed, e.g.,\ local wind conditions, nearby obstacles, and local acoustic and visual requirements for wind turbines. 

\begin{acknowledgements}
This work was supported by the UK Leverhulme Trust under Grant RPG-2013-344.
\end{acknowledgements}

%

\end{document}